\documentclass[preprint,review,letterpaper,12pt,numbers,sort,compress]{elsarticle}

\usepackage{lineno}
\usepackage{hyperref}

\journal{}

\usepackage{amssymb}
\usepackage{amsmath,amssymb,amsfonts, bm}
\usepackage[margin=3.5cm]{geometry}
\usepackage{lipsum}
\usepackage{framed}
\usepackage{academicons}
\usepackage{url}            
\usepackage{booktabs}      
\usepackage{amsfonts}       
\usepackage{nicefrac}       
\usepackage{microtype}      
\usepackage{lipsum}
\usepackage{gensymb}
\usepackage{graphicx}
\usepackage{subfig}
\usepackage[export]{adjustbox}
\usepackage{caption}
\usepackage{physics}
\usepackage{accents}
\usepackage{xcolor}
\usepackage{mathtools}
\usepackage{setspace}
\usepackage{soul}
\usepackage{amsbsy}
\usepackage{tikz}
\usetikzlibrary{tikzmark}
\usetikzlibrary{calc}

\newcommand{\RN}[1]{%
  \textup{\lowercase\expandafter{\romannumeral#1}}%
}
\usepackage{nicematrix,tikz,booktabs}
\usepackage{scalerel}

\DeclareRobustCommand{\redline}{\raisebox{2pt}{\tikz{\draw[-,red,dashed,line width = 1pt](0,0) -- (5mm,0)}}}

\DeclareRobustCommand{\blueline}{\raisebox{2pt}{\tikz{\draw[-,blue,solid,line width = 1pt](0,0) -- (5mm,0)}}}
\definecolor{post_b}{rgb}{0.30,0.75,0.93}
\colorlet{shadecolor}{yellow}

\newcounter{subeq}

\usepackage{algorithm}
\usepackage{algpseudocode}

\newcommand{\orcid}[1]{\href{https://orcid.org/#1}{\includegraphics[width=10pt]{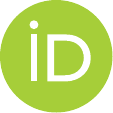}}}
\newtheorem{theorem}{Theorem}

\newcommand{\ubar}[1]{\underaccent{\bar}{#1}}
\begin{document}

\begin{frontmatter}

\title{Gradient Flow Based Phase-Field Modeling Using Separable Neural Networks\\}

\author[label2]{Revanth Mattey \orcid{0000-0002-6732-622X}}
\author[label2,label1]{Susanta Ghosh \orcid{0000-0002-6262-4121}}
\cortext[cor1]{Corresponding author; Email:susantag@mtu.edu}

\address[label2]{Department of Mechanical Engineering--Engineering Mechanics, Michigan Technological University, MI, USA}
\address[label1]{The Center for Data Sciences, The Center for Applied Mathematics and Statistics  Michigan Technological University, MI, USA} 

\begin{abstract}
Allen-Cahn equation is a reaction-diffusion equation and is widely used for modeling phase separation. 
Machine learning methods for solving the Allen-Cahn equation in its strong form suffer from inaccuracies in collocation techniques, errors in computing higher-order spatial derivatives, and the large system size required by the space-time approach.
To overcome these challenges, we propose solving the $L^2$ gradient flow of the Ginzburg–Landau free energy functional, which is equivalent to the Allen-Cahn equation, thereby avoiding the second-order spatial derivatives associated with the Allen-Cahn equation. A minimizing movement scheme is employed to solve the gradient flow problem, eliminating the complexities of a space-time approach.
We utilize a separable neural network that efficiently represents the phase field through low-rank tensor decomposition.
As we use the minimizing movement scheme to numerically solve the gradient flow problem, we thus, refer to the proposed method as the Separable Deep Minimizing Movement (SDMM) method.
The evaluation of the functional in the minimizing movement scheme using the Gauss quadrature technique bypasses the inaccuracies associated with collocation techniques traditionally used to solve partial differential equations.
A hyperbolic tangent transformation is introduced on the phase field prior to the evaluation of the functional to ensure that it remains strictly bounded within the values of the two phases. For this transformation, theoretical guarantee for energy stability of the minimizing movement scheme is established. Our results suggest that this transformation helps to improve the accuracy and efficiency significantly. The proposed method resolves the challenges faced by state-of-the-art machine learning techniques, outperforming them in both accuracy and efficiency. It is also the first machine learning method to achieve an order of magnitude speed improvement over the finite element method. In addition to its formulation and computational implementation, several case studies illustrate the applicability of the proposed method.
\end{abstract}

\begin{keyword}
Gradient flow, Phase field, Separable neural networks, Minimizing movement scheme, Allen-Cahn equation
\end{keyword}

\end{frontmatter}

\section{Introduction}\label{sec:intro}
The phase-field approach is the most widely used method for modeling various microstructure evolution processes. Unlike sharp interface models, phase-field models describe interfaces as a highly localized but smooth transition of phase variables over a finite width, eliminating the need for explicit interface tracking during microstructure evolution \cite{TOURRET2022100810,CHEN2022100868}. 
Two commonly used equations to describe phase-field dynamics are the Cahn-Hilliard nonlinear diffusion equation \cite{baker1969acta} and the Allen-Cahn equation \cite{cahn1977microscopic}. Many diffusion equations, can be interpreted as gradient flows in \( L^2 \) and \( H^{-1} \) spaces respectively \cite{santambrogio2017euclidean,park2023deep}. This work focuses on the Allen-Cahn equation, which is the \( L^2 \)-gradient flow of the Ginzburg–Landau free energy functional \cite{Tonegawa2015,park2023deep}. Gradient flows are fundamental for describing how physical systems evolve toward a state of minimum energy \cite{ambrosio2005gradient}. In general, gradient flows represent the movement of scalar fields such as temperature, concentration, etc. in the steepest descent direction of a free energy functional. The minimizing movement scheme has become a popular method for deriving numerical schemes for solving the gradient flow equations \cite{de1992movimenti,babadjian2014unilateral,muratori2020gradient}. Recently, there has been a rapidly growing interest in developing efficient numerical schemes to solve gradient flow problems \cite{ZHANG2020112743,zaitzeff2020variational}.
In this work we propose a novel machine learning model for the \( L^2 \)-gradient flow of the Ginzburg–Landau free energy functional. In the following, we briefly discuss machine learning approaches for various scientific problems. 

In recent years, there has been a widespread adoption of machine learning (ML) techniques across various fields including image recognition\cite{thiagarajan2021explanation,thiagarajan2022jensen,litjens2017survey,pak2017review}, autonomous mobility\cite{grigorescu2020survey}, and natural language processing\cite{chowdhary2020natural,otter2020survey,nadkarni2011natural,ashish2017attention}, among others. Specifically,  ML techniques like deep neural networks, recurrent neural networks, and graph neural networks have been extensively applied to study computational science and engineering problems over the past decade \cite{kollmannsberger2021deep,frank2020machine,yadav2021interpretable,cuomo2022scientific,pathrudkar2022machine,pathrudkar2024electronic}. Machine learning for solving PDEs has shown remarkable promise and versatility \cite{brunton2023machine,beck2020overview,meuris2023machine}.
Physics-informed neural networks (PINN) is a new class of machine learning framework where the physics information is embedded into the neural networks' loss function \cite{Raissi2019a}. PINNs have found tremendous potential for solving partial differential equations and are widely utilized to solve both forward and inverse problems with a combination of partial physics knowledge and any available data. Their flexibility in discretization and the ability to modify loss functions have led to a wide range of applications \cite{ghaffari2023deep,goswami2022physics,van2022physics,zhang2022analyses,antonelo2024physics,arnold2021state,cai2021physics,song2022versatile}. A much more comprehensive survey of the existing PINN methodologies and their applications are detailed in \cite{cuomo2022scientific, hao2022physics}. In the PINNs framework, several challenges have been identified, including long training times, the need for a large number of collocation points, and difficulties in achieving convergence due to minimizing the residual of the strong form of the PDE. Extensive research has been done in this area to address these challenges in training PINNs for forward problems. Several studies have proposed various techniques such as adaptive sampling based on residual and gradients \cite{wu2023comprehensive,gao2023failure,tang2023pinns}, domain decomposition and sequential learning\cite{mattey2022novel,wight2020solving,diao2023solving,moseley2023finite,jagtap2020extended}, adaptive weighting of the loss function and using adaptive activation functions\cite{wang2022respecting,jagtap2020adaptive}, and modifying the neural network architecture to make it suitable for PINNs \cite{mojgani2023kolmogorov,cho2024separable}. Operator learning is another novel class of machine learning algorithms that aim to learn a specific class of PDEs. Unlike traditional approaches that often require separate training for different PDE instances or classes, operator learning learns the differential operator using a unified framework \cite{li2020fourier,kovachki2021neural,azizzadenesheli2024neural}. The core idea of any operator learning framework is that it uses an integral transform to learn the map between the input and the output space. Deep operator networks (DeepONets) represent a new class of neural network architecture, where two deep neural network are used, one for encoding the input function space, and another neural network for encoding the domain of the output functions \cite{Lu2019DeepONet:Operators,lu2021learning}.

Many of the PINN and operator learning methods found in the literature are centered on reducing the residual of the strong form of the PDE, akin to the collocation technique in numerical methods. However, for highly nonlinear and higher-order PDEs, the collocation method becomes ineffective due to its demand for a large number of points. Minimizing the weak form or the energy functional of a PDE using neural networks has emerged as an efficient approach for obtaining the solution of a PDE \cite{wang2022cenn, kharazmi2019variational}. Notably, minimizing the Ritz energy functional \cite{yu2018deep, liu2023deep, liao2019deep} has shown advantages, particularly in problems with a convex energy landscape. Furthermore, in the context of gradient flows, minimizing an energy-based motion that generalizes the implicit time scheme has been shown to be efficient \cite{park2023deep, li2023phase}.
Separable physics-informed neural networks introduced in \cite{cho2024separable} have shown remarkable speed-up for solving PDEs. The main idea in separable neural networks is that the solution is assumed to be a low-rank tensor approximation. This approximation is constructed by aggregating the tensors predicted at each input dimension using a single multi-layer perceptron (MLP). Furthermore, using forward mode (AD) and leveraging separability drastically reduces the number of collocation points, as each MLP considers a one-dimensional coordinate as its input. \footnote{For example a 3-dimensional system has 3 MLPs for approximating a low-rank tensor in each dimension.}\\

In the present work, at first, we demonstrate that the separable physics-informed neural network is inaccurate for the Allen Cahn equation in the sharp interface limit. To overcome this, we propose a Separable Deep Minimizing Movement scheme, named as SDMM. The key aspects of the proposed SDMM method are:
(1) The proposed approach uses an energy functional that has lower order derivative requirements than the strong form of the Allen-Cahn equation, which alleviates an expensive computational bottleneck.
(2) The minimizing movement scheme simplifies computations by minimizing the functional at each time step, eliminating the complexity of a space-time approach.
(3) The separable neural network models the phase field using a low-rank tensor decomposition, enabling efficient derivative computation through forward-mode automatic differentiation.
(4) The framework allows for the use of Gauss quadrature to accurately calculate the energy functional. 
(5) The use of a hyperbolic tangent transformation of the neural network's output bounds the solution within the values of the two phases. 
(6) The hyperbolic tangent transformation preserves the unconditional energy stability of the minimizing movement scheme. \\

The rest of the paper is organized as follows, in section~(\ref{sec:preliminaries}) basic preliminary concepts of gradient flow, separable neural networks, and its application to the Allen Cahn equation are briefly reviewed; in sections~(\ref{sec:min_mov_scheme}),(\ref{sec:snn_GradientFlow}) first the well known minimizing movement scheme for solving a gradient flow problem is described and next the proposed separable deep minimizing movement (SDMM) scheme is detailed; in section~(\ref{sec:nonlin_transform_boundedness}) the challenges in convergence and boundedness with the proposed method are elaborated and in section~(\ref{sec:tanh_transform}) a nonlinear transformation of the neural network solution to alleviate these challenges is proposed. Furthermore, in section~(\ref{sec:stability_tanh_transform}) a mathematical proof for the unconditional energy stability is also provided; in section~(\ref{sec:num_exp}) three numerical examples have been studied using the proposed method to highlight its advantages; Finally, the conclusions are presented in section~(\ref{sec:conclusion}).

\section{Background}\label{sec:preliminaries}
In this section, some preliminary concepts on gradient flow theory and separable neural networks for approximating phase fields are provided.

\subsection{Gradient Flow for Phase Field Problems}\label{sec:GradientFlow}
This section first introduces $L^2$ gradient flows, which are then utilized to derive the Allen Cahn equation from the Ginzburg-Landau free energy.
\subsubsection{$L^2$ Gradient Flow} \label{sec:L2GradientFlow}
In general, given a curve $\rho$ on a Riemannian manifold ($\mathbb{R}^n$, $g$) and a smooth functional $\mathcal{F}: \mathbb{R}^n \to \mathbb{R}$, then $\rho$ is said to be the gradient flow of $\mathcal{F}$ if it follows the steepest descent direction of $\mathcal{F}$. Mathematically, this can be written as:
\begin{equation}
    \frac{\partial \rho}{\partial t} = -\frac{\delta \mathcal{F}}{\delta \rho}
    \label{eq:gradientFlow}
\end{equation}

\noindent where for any, $\Omega \subset \mathbb{R}^n$, $\frac{\delta \mathcal{F}}{\delta \rho}$ is the functional derivative in the Hilbert space ($L^2 (\Omega)$). Further, let $t \in \mathcal{I} := (0,T] \in \mathbb{R}$ be the time interval, where  $T>0$ is the end time.

\subsubsection{Allen Cahn Equation}\label{sec:AC_Equation}
The Allen-Cahn equation is a $L^2$ Gradient Flow of the Ginzburg-Landau free energy functional. It is a reaction-diffusion equation and is widely used in phase separation problems. For every $\bm{x} \in \mathbb{R}^n$, the Ginzburg-Landau free energy functional is:

\begin{equation}
    \Pi(\phi) = \int_{\Omega} W(\phi) + \frac{\epsilon^2}{2} \abs{\grad \phi}^2 \, d\bm{x}
    \label{eq:freeEnergy}
\end{equation}

\noindent Here, $\phi : (\Omega \times \mathcal{I}) \to [-1,1]$ is the phase field function or an order parameter. The admissable space of the phase field variable $\mathcal{S} := \{\phi \in L^2(\Omega)\, | \, \phi = \ubar{\phi} \,\,\, \text{on} \,\,\, \partial \Omega \}$. $\epsilon$ is a diffuse interface width parameter that is used to parametrize the interface energy density. The first term of the free energy functional, $W(\phi)$, is a double well potential function representing the driving force for phase separation. The following widely used form is considered for the double-well potential, \( W(\phi) = \frac{(\phi^2-1)^2}{4} \). 
Thus, for the Ginzburg-Landau free energy in equation~(\ref{eq:freeEnergy}), the $L^2$ gradient flow is as follows:
\begin{equation}
    \frac{\partial \phi}{\partial t} = -\frac{\delta  \Pi}{\delta \phi}
    \label{eq:GLgradientFlow}
\end{equation}

\noindent Consider the following variations on $\phi$, \qquad
{$\phi \to \phi_{\varepsilon}$} \\
\noindent where, \\
\centerline{$\phi_{\varepsilon} = \phi + \varepsilon\psi\,,  \quad$ 
$\varepsilon \in \mathbb{R}$, $\psi \in \mathcal{V} , \mathcal{V}:= \{\psi \in H^1(\Omega) \, | \, \psi = 0 \,\,\, \text{on} \,\,\, \partial \Omega \}$} 

\noindent The functional derivative is computed by considering a variation of the function from $\phi \to \phi_{\varepsilon}$ where, $\phi_{\varepsilon} = \phi + \varepsilon\psi$. The parameter $\varepsilon \in \mathbb{R}$ is a scalar variable and $\psi$ is the increment to the phase field parameter $\phi$.

\noindent The perturbed functional thus reads as.

\begin{equation}
    \Pi(\phi + \varepsilon \psi) = \int_{\Omega} W(\phi + \varepsilon \psi) + \frac{\epsilon^2}{2} \abs{\grad \phi + \varepsilon \grad \psi}^2 \, d\bm{x}
    \label{eq:perturbed_freeEnergy}
\end{equation}

\noindent After expanding and rearranging the terms in equation~(\ref{eq:perturbed_freeEnergy}), we get 

\begin{equation}
    \Pi(\phi + \varepsilon \psi) =\int_{\Omega} W(\phi) + \frac{\epsilon^2}{2}\abs{\grad \phi}^2 \, d\bm{x} + \varepsilon \int_{\Omega} \frac{\partial W}{\partial \phi} \psi +  \epsilon^2 \grad \phi \cdot \grad \psi \, d\bm{x} + \mathcal{O}(\varepsilon^2)
    \label{eq:rearranged_perturbed_freeEnergy}
\end{equation}

\noindent Using Green's identity and Stokes theorem on equation~(\ref{eq:rearranged_perturbed_freeEnergy}) and neglecting higher order terms, 

\begin{equation}
     \Pi(\phi + \varepsilon \psi) - \Pi(\phi) = \varepsilon \int_{\Omega} \left[\frac{\partial W}{\partial \phi} - \epsilon^2 \grad^2 \phi\right]\psi \, d\bm{x} + \int_{\partial \Omega} \psi \grad \phi \cdot \bm{n} \, d\bm{x}
     \label{eq:simplified_perturbed_freeEnergy}
\end{equation}

\noindent The condition under which the boundary term in equation~(\ref{eq:simplified_perturbed_freeEnergy}) would be zero is: \\
(a) $\grad \phi = 0 \rightarrow $ Natural Boundary condition (no flux condition) or \\
(b) $\psi = 0 \rightarrow $ when essential boundary conditions for $\phi$ are prescribed on $\partial \Omega$. \\

\begin{equation}
    \Pi(\phi + \varepsilon \psi) - \Pi(\phi) = \varepsilon \int_{\Omega} \left[\frac{\partial W}{\partial \phi} - \epsilon^2 \grad^2 \phi\right]\psi \, d\bm{x}
    \label{eq:final_perturbed_freeEnergy}
\end{equation}

\noindent It can be observed that equation~(\ref{eq:final_perturbed_freeEnergy}) is the Gateaux derivative of the energy functional ($\Pi$) in the direction of ($\psi$),

\begin{equation}
    \left\langle \frac{\delta \Pi}{\delta \phi},\psi \right\rangle_2 := \lim_{\varepsilon \to 0} \frac{\Pi(\phi + \varepsilon \psi) - \Pi(\phi)}{\varepsilon} \\
    =\left\langle \frac{\partial W}{\partial \phi} - \epsilon^2 \grad^2\phi,\psi \right\rangle_2
    \label{eq:DirectionalDeriative}
\end{equation}

\noindent Here, $\langle \, , \, \rangle_2$ denotes the $L^2$ inner product over $\Omega$. Using the result from equation~(\ref{eq:DirectionalDeriative}) and substituting the functional derivative ($\frac{\delta \Pi}{\delta \phi}$) in equation~(\ref{eq:GLgradientFlow}) yields the Allen Cahn equation as follows,
\begin{equation}
    \frac{\partial \phi}{\partial t} =  \epsilon^2 \grad^2\phi - \frac{\partial W}{\partial \phi}
    \label{eq:AC_Equation}
\end{equation}
Where, $\grad^2$ follows the standard definition as $\grad \cdot \grad$


\subsection{Phase Field Modeling using Separable Neural Networks}\label{sec:backgroundSPINN}
State-of-the-art machine learning models called Separable Physics Informed Neural Networks (SPINN) are utilized to solve partial differential equations \cite{cho2024separable}. In SPINN, the assumption is made that the output of the neural network is a low-rank tensor approximation of individual rank-1 tensors across all dimensions. By leveraging the concepts of forward automatic differentiation (\ref{sec:AD}) and the separability of functions (\ref{sec:sep_vs_nonsep}), SPINN has demonstrated remarkable speed-up in solving PDEs. However, SPINN faces challenges in solving sharp interface phase field problems, such as the Allen Cahn equation. This section first introduces separable neural networks and then highlights the challenges faced in solving sharp interface phase field problems. \\

\subsubsection{Separable Neural Networks}\label{sec:snn}
Separable neural networks (SNN) are a new class of neural networks where the solution is approximated by considering a single MLP (multi-layer perceptron) for each dimension of the system \cite{cho2024separable}. In a single SNN, $d$ MLPs are considered, where $d$ is the number of dimensions in the system. Each MLP then takes a one-dimensional coordinate component as input and predicts a $m$-dimensional feature representation, $g^{(\alpha_i)} : \mathbb{R} \to \mathbb{R}^m$. The solution is then aggregated at each grid point by taking the inner product of the feature vectors predicted across the $d$ dimensions. This can be written as:
\begin{equation}
    \Tilde{\phi} \approx \sum_{j=1}^{m} \prod_{i=1}^{d} g^{(\alpha_i)}_j (x_i)
    \label{eq:spinn_sol}
\end{equation}

\noindent Where, $(x_1, x_2, \ldots, x_d)$ are the co-ordinates of a grid point for a $d$ dimensional system and $\Tilde{\phi}$ is the output of the separable neural networks. 

\subsubsection{Separable Physics Informed Neural Networks (SPINN)} 
In the physics-informed neural networks (PINNs) framework, boundary value problems are solved by encoding it in a neural network's loss function. 
The commonly used PINNs approaches use the \textit{strong form} of the PDE to obtain the solution by minimizing the residual of the PDE predicted using the neural network's solution.  
The physics-informed loss consists of three components, (1) error in the initial condition, (2) error in the boundary condition, and (3) residual error in the partial differential equation. The residual term in the loss function ensures that the predicted solution satisfies the underlying boundary value problem. \\  

\noindent Consider a general partial differential equation denoted by, 
\begin{equation*}
    \mathcal{N}(\bm{x}, \phi, \grad \phi,\grad^2\phi, \cdots, \grad^q \phi) = 0;  \quad  \bm{x} \in \Omega \subset \mathbb{R}^n
\end{equation*}
Where, $\partial \Omega \in \mathbb{R}^n$ denotes the boundary of $\Omega$ and  $(0,\,T] = \mathcal{I} \subset \mathbb{R}$ denotes the time domain. For the given system, the initial condition and the boundary condition are given by $\phi{(\bm{x},0)}=\phi_0{(\bm{x})},\,\bm{x}\in \Omega$ and $\phi{(\bm{x},t)}=\phi_{bc}{(\bm{x},t)},\,\bm{x}\in \partial\Omega,\, t\in \mathcal{I}$ respectively. The coordinates where the residual of the PDE, boundary conditions, and initial conditions are minimized are represented as $(\bm{x}^r_k,t^r_k),(\bm{x}^{b}_k,t^b_k),(\bm{x}^i_k,0)$ respectively. The neural network approximation of the solution is given by $\Tilde{\phi}(\bm{x},t)$. The physics informed loss $\mathcal{L}_{PINN}$ reads as:

\begin{equation}
  \mathcal{L}_{PINN} =  \mathcal{L}_{IC} + \mathcal{L}_{BC} + \mathcal{L}_{PDE}
    \label{eq:pinn_loss}
\end{equation}

\begin{align}
    \mathcal{L}_{PDE} =  \frac{1}{N_r}\sum_{k=1}^{N_r}\left(\mathcal{N}[\Tilde{\phi}](\bm{x}^r_k,t^r_k) \right)^2 \label{eq:pinn_loss_res}\\
    \mathcal{L}_{BC}  =  \frac{1}{N_b}\sum_{k=1}^{N_b}\left(\Tilde{\phi}(\bm{x}_k^b,t_k^b) - \phi_{bc}(\bm{x}_k^b,t_k^b)\right)^2  \label{eq:pinn_loss_bc} \\
    \mathcal{L}_{IC}  =  \frac{1}{N_i}\sum_{k=1}^{N_i}\left(\Tilde{\phi}(\bm{x}^i_k,0) - \phi_0(\bm{x}^i_k)\right)^2
    \label{eq:pinn_loss_ic}
\end{align}

\noindent Given the standard physics-informed loss function, a separable neural network has been trained to solve a sharp interface problem, as discussed in the following section.

\subsubsection{Challenges of SPINN for Sharp Interface Phase Field Evolution}\label{sec:spinn_challenges}
In this section, a 1D time-varying Allen Cahn equation is solved using the concept of separable neural networks. A physics-based loss function given in equations~(\ref{eq:pinn_loss}, \ref{eq:pinn_loss_res}, \ref{eq:pinn_loss_bc}, \ref{eq:pinn_loss_ic}) where the strong form of the PDE is solved. The details of the boundary value problem are given below:

\begin{align}
\centering
    & \frac{\partial \phi}{\partial t} =  \epsilon^2 \grad^2\phi - f(\phi) \\
    & \epsilon = 0.01, f(\phi) = 5(\phi^3 - \phi) \\
    & \phi(x,0) = x^2 \cos{(\pi x)}
\end{align}

where the domain, $\Omega \times T = [-1,1] \times (0,1]$. Here, for the given system periodic boundary conditions are considered. The neural network training involves utilizing a mesh grid comprising 256 points in each spatial and temporal direction. This results in a total of 65,536 collocation points where the partial differential equation (PDE) residual ($N_r$) is minimized. Additionally, 256 points are considered for the initial condition ($N_i$), while 512 points are used for the boundary condition ($N_b$). \\

\begin{figure}[b!]
    \centering
    \includegraphics[scale = 0.3]{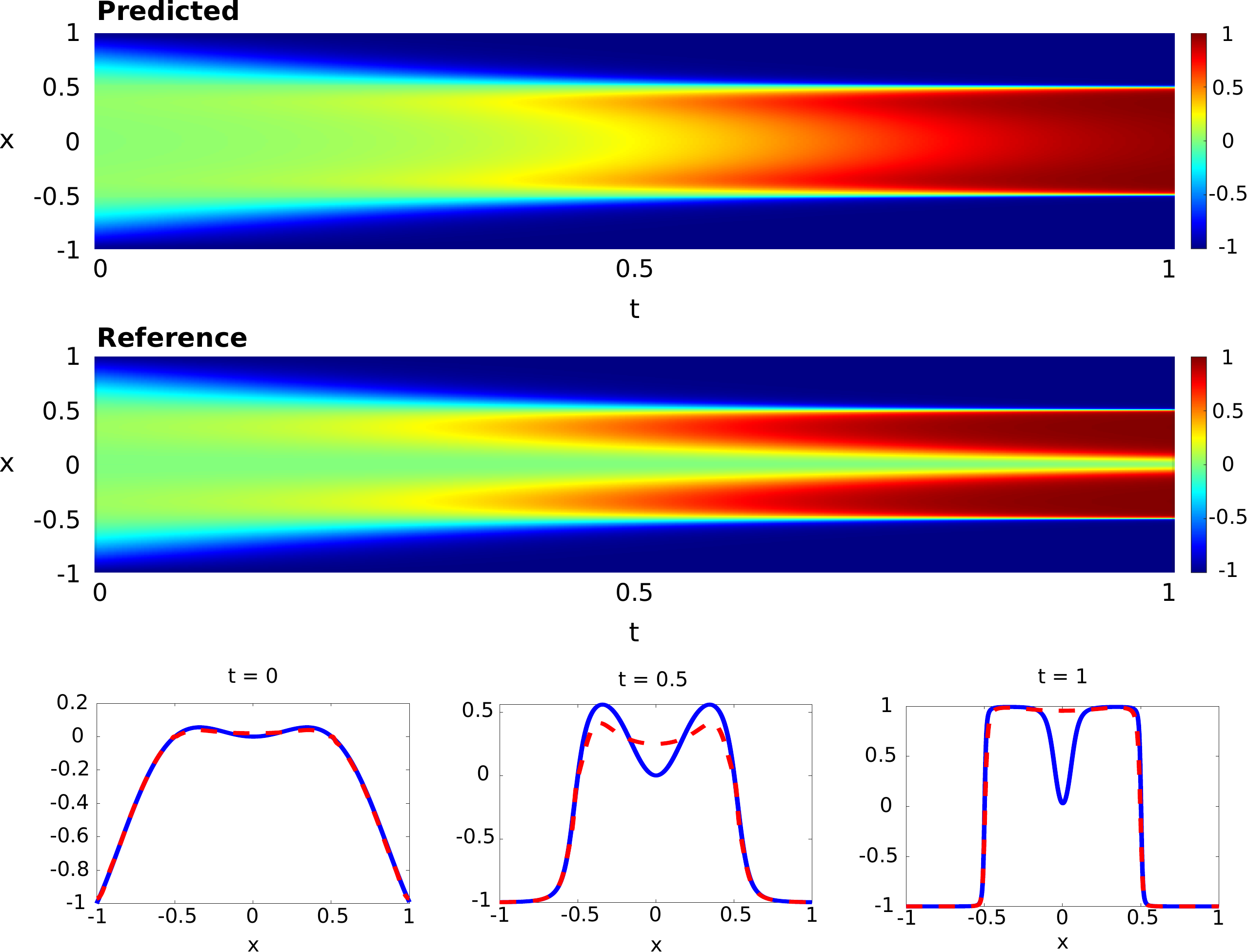}
    \caption{(Top, Middle) Space-time phase filed contour by the SPINN method and the reference solution. (Bottom) Solution at various time obtained by the SPINN method (\blueline) and the reference solution (\redline) obtained via Chebfun \cite{driscoll2014chebfun}}
    \label{fig:AC_1D_pred_exact}
\end{figure}

\noindent As shown in figure~(\ref{fig:AC_1D_pred_exact}), the SPINN approach faces challenge in approximating the solution at sharp interface jumps (for small $\epsilon$ values). Consequently, to address this issue, a moving minimization scheme described in the subsequent text is employed to sequentially obtain the solution.

\section{Proposed Method: Separable Deep Minimizing Movement (SDMM) Scheme}\label{sec:ProposedMethod}

In the current section, the standard minimizing movement scheme and energy stability are discussed. Then, the proposed separable neural network for approximating the phase field and details of the loss function and its computation are detailed. Finally, a novel non-linear transformation that maintains energy stability, keeps the solution bounded and improves accuracy is demonstrated.

\subsection{Minimizing Movement Scheme}\label{sec:min_mov_scheme}
The minimizing movement scheme \cite{de1992movimenti,de1993new} is a generalized method for the study of the steepest descent curves of a functional in a metric space. It is an energetic formulation of the implicit backward Euler scheme that describes the evolution of the phase field function ($\phi$). Given the Allen Cahn equation in equation~(\ref{eq:AC_Equation}), the implicit backward Euler method for obtaining the solution at time $t + \tau$ given the solution at time $t$ and a time step $\tau \, > \, 0$ is,

\begin{equation}
    \frac{\phi_{t+\tau} - \phi_t}{\tau} = \left. \epsilon^2 \grad^2 \phi_{t+\tau} - \frac{\partial W}{\partial \phi} \right\vert_{\phi_{t+\tau}}
    \label{eq:euler_discretization_AC}
\end{equation}

\noindent This implicit Euler equation is the first-order optimality condition of the following optimization problem,

\begin{equation}
    \phi_{t+\tau} = \underset{\phi}{\mathrm{argmin}} \left(\Pi(\phi) + \frac{1}{2\tau} \norm{\phi - \phi_t}^2\right)
    \label{eq:mov_min}
\end{equation}

\noindent Where, $\norm{\bullet}^2 = \langle \bullet, \bullet\rangle_2$. Equation~(\ref{eq:mov_min}) provides a means to obtain the dynamic evolution of the energy functional $\Pi(\phi)$ by solving a stationary optimization problem at each time step. Additionally, the quadratic term $\frac{1}{2\tau}\norm{\phi - \phi_t}^2$ is commonly referred to as the \textit{movement limiting} term counteracts the deviation of the solution from the current configuration.

\subsubsection{Unconditional Stability of Minimizing Movement Scheme}
The time discretization obtained through the minimizing movement scheme is considered as unconditionally stable if 
\begin{equation}\label{eq:UnconditionalStability}
    \Pi(\phi_{t+\tau}) \leq \Pi(\phi_{t}),\qquad \forall t \in [0,T]
\end{equation}


\subsection{Proposed Method: Separable Neural Networks Based Minimizing Movement Scheme for Phase Field Problems} \label{sec:snn_GradientFlow}
Gradient flow problems at sharp interface limits are mathematical models that describe how interfaces between different phases within a system evolve. These interfaces are usually characterized by steep gradients and sudden transitions. The presence of such sharp gradients at these interfaces poses a challenge for neural networks when trying to predict the solution using only the residual of the strong form of a PDE. Moreover, this approach can result in inaccuracies and errors in model predictions, particularly near the interfaces where the gradients are most significant. \\

\noindent In this work, the moving minimization scheme described in section~(\ref{sec:min_mov_scheme}) has been adopted to solve the sharp interface gradient flow problem. Effectively, this is an iterative approach where the solution is obtained by minimizing the energy term and movement term simultaneously. The main advantage of such an approach is that instead of minimizing the residual of the PDE, which requires a large number of collocation points in space and time, the free energy at every time step is minimized. \\

\noindent Thus, to approximate the solution only in the spatial dimensions a separable neural network described in section~(\ref{sec:snn}) has been utilized. 
The solution is then sequentially obtained by minimizing the functional described in equation~(\ref{eq:mov_min}) using the neural network's approximation. Through utilization of the moving minimization scheme alongside separable neural networks makes our methodology exhibit similarities to a time-stepping algorithm. In the next section, the details of the loss function and the algorithmic approach to obtain the solution are presented. \\

\noindent The notation scheme used for the rest of the paper remains unaltered from those described in sections~(\ref{sec:GradientFlow})-(\ref{sec:min_mov_scheme}).

\subsubsection{Loss Function}\label{sec:snn_loss}
As described in sections~(\ref{sec:min_mov_scheme}),(\ref{sec:snn_GradientFlow}), the solution for a gradient flow problem is obtained by sequentially minimizing the sum of energy loss and the movement loss. The total loss function to be minimized is denoted by $\mathcal{L}_{SDMM}$ and takes the following form,
\begin{equation}
\mathcal{L}_{SDMM} = \mathcal{L}_{\mathrm{E}} + \mathcal{L}_{\mathrm{M}}
\label{eq:snn_loss_movmin}
\end{equation}

\noindent Where, the energy loss $\mathcal{L}_{\mathrm{E}}$ and movement loss $\mathcal{L}_{\mathrm{M}}$ are given by,

\begin{align}
    \mathcal{L}_{\mathrm{E}} &= \int_{\Omega} W(\Tilde{\phi}) + \frac{\epsilon^2}{2} \abs{\grad \Tilde{\phi}}^2 \, d\bm{x}\\
    \mathcal{L}_{\mathrm{M}} &= \frac{1}{2\tau}\norm{\Tilde{\phi} - \Tilde{\phi_t}}^2
\end{align}

\noindent To simplify notation, let's denote a time sequence, ($t = 0,\tau,2\tau,\cdots,N_t\,\tau$),  where the solution needs to be computed. Here, $N_t$ is the total number of time intervals such that $N_t\,\tau=T$. At first, given the initial condition $\phi{(\bm{x},0)}=\phi_0{(\bm{x})},\,\bm{x}\in \Omega$, the solution at time $\tau$ is obtained by minimizing the energy loss ($\mathcal{L}_E$) along with the movement loss ($\mathcal{L}_M$). The movement loss ($\mathcal{L}_M$) minimizes the difference between the neural network output $\Tilde{\phi}$ and $\phi_0$. Subsequently, the same process is repeated to obtain the solution until time $T$. The next section presents the details regarding evaluating gradients and integrals in a neural network framework. 

\subsubsection{Numerical Integration}\label{sec:num_integration}
In this section, we provide the details of the numerical integration scheme used to compute the above loss function.  Let the domain $\Omega$ be partitioned into a collection of elements denoted by $\mathcal{T}$, where \\

$\mathcal{T} = \left\{\Omega_e : \Omega_e \, \text{ is an element(sub-domain) of } \, \Omega\right\}$, such that, 
$\Omega = \bigcup_{\Omega_e \in \mathcal{T}} \, \Omega_e$ \\

In the current work, Gaussian quadrature is employed to evaluate the integrals. With this chosen integration method, the required derivatives are computed at the quadrature points using the automatic differentiation technique described in section~(\ref{sec:AD}). For any general function $h(\bm{x})$, the Gaussian quadrature rule over an element sub-domain $\Omega_e$ is given by,

\begin{equation}
    \int_{\Omega_e} h(\bm{x}) \; d\bm{x} \approx |\mathcal{J}| \sum_{j=1}^{m} w_j h(\bm{x}^g_j) \; 
\end{equation}

\noindent where, $\bm{x}^g_j$ are the gauss points (abscissas) in the element $\Omega_e$ and $w_j$ are the weights. The determinant $|\mathcal{J}|$ represents the determinant of the Jacobian matrix that describes the mapping between the standard parent element\footnote{Standard parent element is a quadrilateral with domain $[-1,1]\times[-1,1]$} and the mapped element. In the current study, a 4-point Gauss quadrature method has been utilized for regular quadrilateral elements for two-dimensional problems.



\subsection{A Nonlinear Transformation of Phase Field to Ensure Boundedness of the Solution} \label{sec:nonlin_transform_boundedness} 
The free energy functional given in equation~(\ref{eq:freeEnergy}), corresponding to the Allen Cahn equation (\ref{eq:AC_Equation}), is solved using the proposed method. We have observed that the solution exhibits undulations at locations away from the phase boundaries, resulting in phase values ($\phi$) slightly beyond the range $[-1,1]$. These undulations might be attributed to the limited number of optimization iterations in each time step or to a smaller network size. 
We have noticed that increasing neural network size and/or the number of iterations in the training of the neural network at each time step improves the result but can not completely alleviate the problem. Additional details on this issue is provided in section~\ref{sec:coarsening})

\subsubsection{`\,$\tanh$'  Transformation of Phase Field in the Proposed SDMM Approach} \label{sec:tanh_transform}
To mitigate the aforementioned issue faced by the proposed method, a nonlinear transformation is employed on the phase field. In particular, a `$\tanh$' mapping is proposed that ensures that the solution lies within the range $[-1,1]$, as
\begin{equation}
    \Tilde{\phi} \mapsto \tanh{\Tilde{\phi}} \,=\,\phi
\end{equation}
Where $\Tilde{\phi}$ is the neural network's output. 
The phase field, $\phi$ obtained by transforming $\Tilde{\phi}$ is used in the loss function mentioned in equation~\ref{eq:snn_loss_movmin}. 
This choice aligns with the physical constraints of the problem, as the phase should not realistically exceed these bounds. Moreover, using the `$\tanh$' transformation has shown improved convergence rates, as evidenced by a comparative benchmarking study presented in the results section (\ref{sec:coarsening}) for a coarsening problem. 
In the following, the unconditional energy stability of the proposed (`$\tanh$' transformed) SDMM approach is analyzed.

\subsubsection{Unconditional Stability of Proposed `$\tanh$'  Transformed  SDMM Approach} \label{sec:stability_tanh_transform} 
\begin{theorem}
Let a phase field, $\Tilde{\phi}$, is transformed to $\Tilde{\phi} \mapsto \tanh{\Tilde{\phi}} \,=\,\phi$, then the minimizing movement scheme for the Ginzburg-Landau functional on the transformed phase field $\phi$ is unconditionally stable, i.e., satisfies the equation~(\ref{eq:UnconditionalStability}).
\end{theorem}
\noindent \textbf{Proof.} The rate of change of the  Ginzburg-Landau energy functional is given by,
\begin{equation}
    \frac{d\Pi(\phi)}{dt} = \int_{\Omega} \frac{\delta \Pi(\phi)}{\delta \phi} \cdot \frac{d\phi}{dt} d\bm{x} = \left\langle \frac{\delta \Pi(\phi)}{\delta \phi}, \frac{d\phi}{dt}\right\rangle_2
    \label{eq:timeDerivative_Energy}
\end{equation}
\noindent Applying the transformation  $\Tilde{\phi} \mapsto \tanh{\Tilde{\phi}} \,=\,\phi$,

\begin{equation}
    \frac{d\Pi(\tanh{\Tilde{\phi}})}{dt} = \int_{\Omega} \frac{\delta \Pi (\tanh{\Tilde{\phi}})}{\delta \tanh{\Tilde{\phi}}} \cdot \frac{d\tanh{\Tilde{\phi}}}{dt} d\bm{x}
    \label{eq:transformed_timeDerivative_Energy}
\end{equation}
\noindent We obtain the expression for the first term of the integrand of the right-hand side as, 
\begin{equation}
    \frac{\delta \Pi (\tanh{\Tilde{\phi}})}{\delta \tanh{\Tilde{\phi}}} = \frac{1}{\epsilon^2} \frac{\partial W(\tanh{\Tilde{\phi}})}{\partial \tanh{\Tilde{\phi}}} + 2\sech^2{\Tilde{\phi}}\tanh{\Tilde{\phi}}\,|\grad \Tilde{\phi}|^2 - \sech^2{\Tilde{\phi}}\grad^2\Tilde{\phi} 
    \label{eq:variationalDerivative_energy_tanh}
\end{equation}
\noindent Consider the augmented functional of the minimizing movement scheme,  i.e the sum of the energy functional and the movement limiting term, given in the right-hand side of equation~(\ref{eq:mov_min}). 
Let us apply the transformation $\Tilde{\phi} \mapsto \tanh{\Tilde{\phi}} \,=\,\phi$ on the augmented functional and denote it by $\mathcal{G}$, as 
\begin{equation}
    \mathcal{G} = \Pi(\tanh \Tilde{\phi}) + \frac{1}{2\tau} \left[\tanh{\Tilde{\phi}} - \tanh{\Tilde{\phi_{t}}} \right]^2
\end{equation}
Using the Ginzburg-Landau functional yields, 
\begin{equation}
    \mathcal{G} = \frac{1}{2}|\grad \tanh{\Tilde{\phi}}|^2 + \frac{1}{\epsilon^2} W(\tanh{\Tilde{\phi}}) + \frac{1}{2\tau} \left[\tanh{\Tilde{\phi}} - \tanh{\Tilde{\phi_{t}}} \right]^2
    \label{eq:minmov_tanh}
\end{equation}

\noindent The variation of the functional $\mathcal{G}$ with respect to $\Tilde{\phi}$ can be computed as follows
\begin{equation}
    \frac{\delta\mathcal{G}}{\delta \Tilde{\phi}} = \frac{\partial \mathcal{G}}{\partial \Tilde{\phi}} - \sum_{j=1}^n \frac{\partial}{\partial x_j}\left(\frac{\partial \mathcal{G}}{\partial g_j}\right),\\
    \text{where,}\, g_j = \frac{\partial\mathcal{G}}{\partial x_j}
    \label{eq:functionalDerivative}
\end{equation}
\noindent where $n$ is the number of dimensions in the domain.
\noindent Substituting the functional $\mathcal{G}$ in equation~(\ref{eq:functionalDerivative}) yields,
\begin{multline}
    \frac{\delta\mathcal{G}}{\delta \Tilde{\phi}} = 2\sech^4{\Tilde{\phi}}\tanh{\Tilde{\phi}}\,|\grad \Tilde{\phi}|^2 + \frac{1}{\epsilon^2} \frac{\partial W(\tanh{\Tilde{\phi}})}{\partial \tanh{\Tilde{\phi}}}\sech^2{\Tilde{\phi}} \\+ \frac{1}{\tau} \left[\tanh{\Tilde{\phi}} - \tanh{\Tilde{\phi_{t}}} \right]\sech^2{\Tilde{\phi}}  -\sech^4{\Tilde{\phi}}\grad^2\Tilde{\phi}
    \label{eq:minmov_variationalDerivative}
\end{multline}
\noindent Using the first order optimality condition, i.e. equating the variational derivative $\left(\frac{\delta\mathcal{G}}{\delta \Tilde{\phi}}\right)$ to zero yields the $\Tilde{\phi}$ at time ${t+\tau}$ according to the minimizing movement scheme. Rearranging the above equation, 
\begin{multline}
    \frac{\tanh{\Tilde{\phi}_{t+\tau}} - \tanh{\Tilde{\phi_{t}}}}{\tau} = \left. \sech^2{\Tilde{\phi}}\grad^2\Tilde{\phi} -2\sech^2{\Tilde{\phi}}\tanh{\Tilde{\phi}}\,|\grad \Tilde{\phi}|^2 - \frac{1}{\epsilon^2} \frac{\partial W(\tanh{\Tilde{\phi}})}{\partial \tanh{\Tilde{\phi}}} \, \right\vert_{t+\tau}
    \label{eq:dtanh_dtau}
\end{multline}
\noindent From equations~(\ref{eq:variationalDerivative_energy_tanh}) and (\ref{eq:dtanh_dtau}), we get
\begin{equation}
    \frac{\tanh{\Tilde{\phi}_{t+\tau}} - \tanh{\Tilde{\phi_{t}}}}{\tau}  
    \approx  \frac{d\tanh{\Tilde{\phi}}}{dt}\, 
    = \left. -\frac{\delta \Pi (\tanh{\Tilde{\phi}})}{\delta \tanh{\Tilde{\phi}}}\, \right\vert_{t+\tau}
    \label{eq:variationalEnergyDerivative_dtanhdtau}
\end{equation}
\noindent Finally, substituting the result obtained in equation~(\ref{eq:variationalEnergyDerivative_dtanhdtau}) in (\ref{eq:transformed_timeDerivative_Energy}) yields the following at time $t+\tau$, 
\begin{equation}
    \frac{d\Pi(\tanh{\Tilde{\phi}})}{dt} =  \left\langle  \frac{\delta \Pi (\tanh{\Tilde{\phi}})}{\delta \tanh{\Tilde{\phi}}} , -\frac{\delta \Pi (\tanh{\Tilde{\phi}})}{\delta \tanh{\Tilde{\phi}}}\right\rangle_2 \,  = -\,\norm{\frac{\delta \Pi (\tanh{\Tilde{\phi}})}{\delta \tanh{\Tilde{\phi}}}}^2\; \leq 0 \label{eq:transformed_timeDerivative_Energy_decreasing}
\end{equation}
\noindent Hence it is proved that under the transformation, $\Tilde{\phi} \mapsto \tanh{\Tilde{\phi}} \,=\,\phi$, the minimizing movement scheme for the free energy functional ($\Pi$) remains unconditionally energy stable. \\

\noindent The `$tanh$' transformation allows the prediction of the separable neural network $\Tilde{\phi}$ to go beyond the range $[-1,1]$ that defines two phases, while the $\phi$ remains within that range. From this point forward, the SDMM approach using the `$\tanh$' transformation will be referred to simply as SDMM, unless stated otherwise.

\section{Numerical Experiments}\label{sec:num_exp}
In this section, details are first provided about the separable neural network used in the proposed SDMM method, the computation of the reference solution using the Finite Element method, and the error metrics used to compute the error in the proposed SDMM method. 
Following this, three numerical experiments are presented, in which the gradient flow of Ginzburg–Landau free energy functional given in equation~(\ref{eq:freeEnergy}) is solved for three different initial conditions. In all the examples, a double-well potential, $W(\phi) = \frac{(\phi^2-1)^2}{4}$, is considered, and the interfacial thickness parameter $\epsilon$ has been set to 0.01. In the following, $\tau = t_{n+1}-t_{n}$ denotes the time step. 

\subsection{Details of the Separable Neural Network }\label{sec:nn_details}
The architecture of the separable neural network described in section~(\ref{sec:snn}) consists of `$d$\,' multi-layer perceptrons (MLPs) for solving a $d$-dimensional system. In the current study, two-dimensional time-varying PDEs are considered. Thus, two MLPs are required for approximating the solution field, where each MLP consists of  4 hidden layers with 128 neurons and an output layer with 256 neurons. Gaussian linear unit (commonly referred to as GELU) is chosen as the activation function. 
In the proposed SDMM method, at the time $t=0$,  the given initial condition is learned by minimizing the error in the initial condition with the ADAM optimizer. For the subsequent time steps, the LBFGS optimizer is utilized to minimize the loss function given in equation~(\ref{eq:snn_loss_movmin}).

\subsection{Reference Solution}\label{sec:ref_sol}
To obtain the reference solution for the subsequent numerical experiments, the Finite Element method (FEM) is used on a very refined spatial and temporal grid until a convergence is achieved. The Finite Element solutions are obtained using the FEniCS simulation package \cite{ScroggsEtal2022,BarattaEtal2023}. FEniCS is a widely utilized open-source computing platform for solving partial differential equations (PDEs) using the FEM. To solve the time-dependent PDEs with FEniCS an explicit scheme is utilized to discretize the time derivative. The reference solutions obtained using FEniCS serve as benchmarks for evaluating the accuracy of the proposed SDMM method in the numerical experiments described below.

\subsection{Error Metrics}\label{sec:err_sol}
The $L^2$ norm of the difference between the SDMM and the reference solutions is used to compute the error in the SDMM method as,
\begin{equation}
    \mathcal{E}_{\text{SDMM}} = \frac{1}{N_t}\sum_{k=1}^{N_t} \sqrt{\norm{\phi(\bm{x},t_k) - \phi_{ref}(\bm{x},t_k)}^2}
    \label{eq:err_rmse}
\end{equation}
Here, $t_k$ denotes the $k$-th time step.
\noindent Here, $\phi(\bm{x},t_k)$ is the solution predicted by the proposed SDMM approach and $\phi_{ref}(\bm{x},t_k)$ is the reference solution at the $k$-th time step.  
Furthermore, to compare the difference between the phase fields obtained by the SDMM and the reference solution, the absolute error used ($ \phi_{\text{abs-error}}$)is utilized.
\begin{equation}
    \phi_{\text{abs-error}} = \abs{\phi - \phi_{ref}}
    \label{eq:err_abs}
\end{equation}

\subsection{Test 1: Star-Shaped Interface Problem}\label{sec:star_int}
\begin{figure}[H]
    \centering
    \includegraphics[scale=0.375]{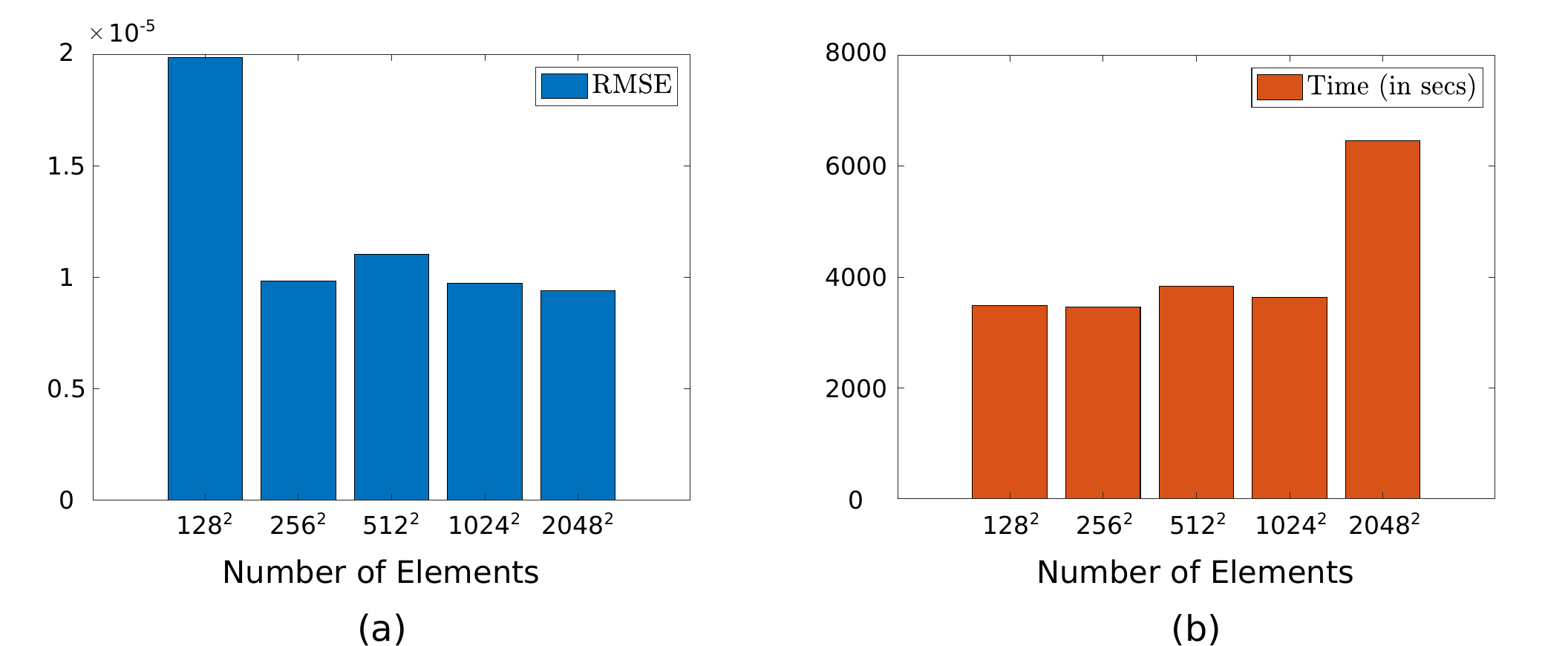}
    \caption{(a) Errors ($\mathcal{E}_{\text{SDMM}}$) in the SDMM method with respect to the reference solution, (b) Simulation (wall-clock) times for different mesh sizes. We found that the SDMM method provides erroneous solutions for mesh size coarser than $128^2$.} 
    \label{fig:ic3_err_mesh_time}
\end{figure}

The evolution of a star-shaped interface is a well-studied problem of the Allen-Cahn equation as it has curvature-driven dynamics. The computational domain is chosen as $\Omega = [0,1]\times[0,1] \in \mathbb{R}^2$. The value of the time step is taken as $\tau = 2 \times 10^{-5}$ and is simulated until a total time of $T = 0.02$. The initial condition is a radially symmetric star-shaped function, centered at ($0.5,0.5$), and is given by
\begin{equation}
    \phi_0(x,y) = \tanh{\frac{R_0 + 0.1 \cos{7\theta} - \sqrt{(x-0.5)^2 + (y-0.5)^2}}{\sqrt{2}\epsilon}}
    \label{eq:ic_star}
\end{equation}

\noindent where, $R_0 = 0.25$ and $\theta$ varies across the grid as follows:

\begin{equation}
    \theta = \begin{cases}
        \tan^{-1}\left(\frac{y-0.5}{x-0.5}\right), & x > 0.5 \\ 
        \pi + \tan^{-1}\left(\frac{y-0.5}{x-0.5}\right), & \text{otherwise}
    \end{cases}
    \label{eq:ic_star_theta}
\end{equation}
\noindent 
As mentioned in section~(\ref{sec:AC_Equation}), no-flux boundary condition ($\grad \phi = 0$) has been considered in this work. 
The reference solution is obtained using FEniCS on a  $2048 \times 2048$ finite element mesh. The wall time for this simulation is $56,771$ seconds on 24 CPU cores in parallel (see Appendix~\ref{sec:comp_resources} for more details on the computer). 
To obtain the errors in the SDMM solutions that are computed on various meshes, the SDMM solution is predicted on a mesh that contains the same number of nodes as the reference solution. The errors and the computational times required in the SDMM solutions for various mesh sizes are given in Table~\ref{tab:rel_error_IC3} and figure~\ref{fig:ic3_err_mesh_time}.
\begin{table}[H]
    \centering
    \begin{tabular}{c|c|c}
        $\mathcal{T}$ (No. of Elements)  &  $\mathcal{E}_{\text{SDMM}}$ & Computational time (in secs)\\
        \toprule
         $128^2$ & $1.9860 \times 10^{-5}$ & 3490.60\\
         $256^2$ & $9.8355 \times 10^{-6}$  & 3460.72\\
         $512^2$ & $1.1036 \times 10^{-5}$ & 3834.25\\
         $1024^2$ & $9.7467 \times 10^{-6}$  & 3628.80\\
         $2048^2$ & $9.3992 \times 10^{-6}$  & 6465.70\\
        \bottomrule
    \end{tabular}
    \caption{Errors in the SDMM solution relative to the reference solution, along with the corresponding computational times for various mesh sizes, are provided.}
    \label{tab:rel_error_IC3}
\end{table}

\begin{figure}[H]
    \centering
    \includegraphics[scale=0.425]{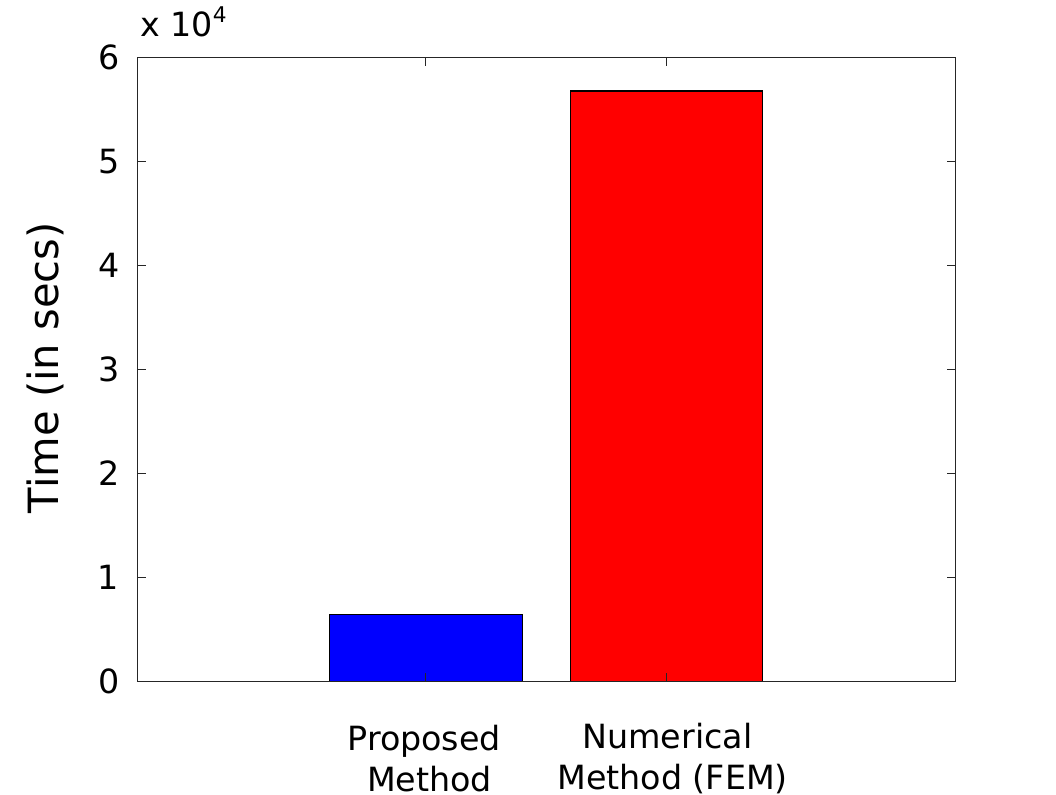}
    \caption{Bar graph comparing the computational wall time for the proposed SDMM method using GPU and the numerical method employing 24 CPU cores. Details of the CPU and GPU systems used for the simulations are given in \ref{sec:comp_resources}.}
    \label{fig:ic3_refvsNN_time}
\end{figure}

\begin{figure}[H]
    \centering
    \includegraphics[scale=0.5]{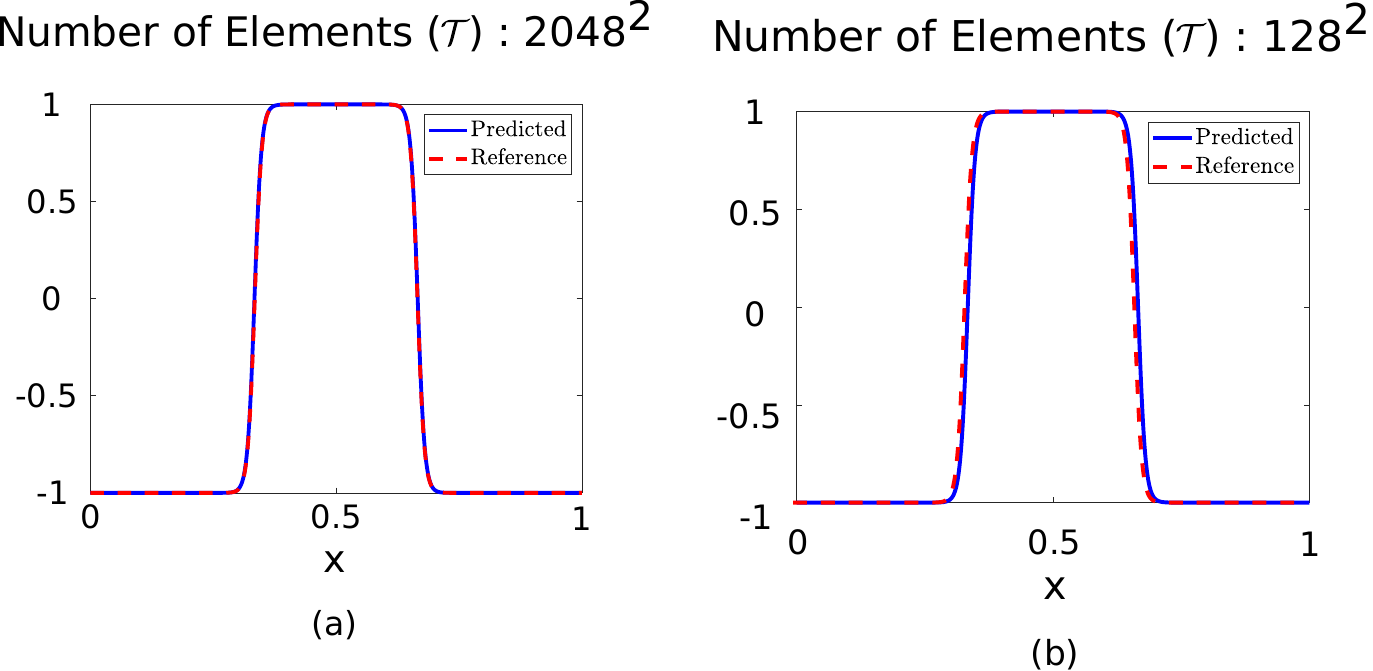}
    \caption{Cross section of the solution predicted at time, t = 0.02 secs with number of elements (a) $2048^2$ (b) $128^2$.}
    \label{fig:ic3_comparison_n128-n2048}
\end{figure}

\noindent From figure~(\ref{fig:ic3_err_mesh_time}) it can be observed that refining the mesh until $1024^2$ elements does not increase the computational time using the SDMM method significantly, however for the $2048^2$ elements the computational time increase significantly. Mesh refinement reduces the error $\mathcal{E}_{\text{SDMM}}$ in the SDMM method significantly until the mesh $256^2$, however, beyond that further mesh refinement does not improve the accuracy significantly. Figure~(\ref{fig:ic3_refvsNN_time}) shows a comparison of time required to obtain the solution using the proposed SDMM approach and the finite element method. Further, as depicted in figure~(\ref{fig:ic3_comparison_n128-n2048}), employing a finer mesh with $2048^2$ elements accurately captures the sharp jump compared to a coarser mesh with $128^2$ elements.

\begin{figure}[H]
    \centering
    \includegraphics{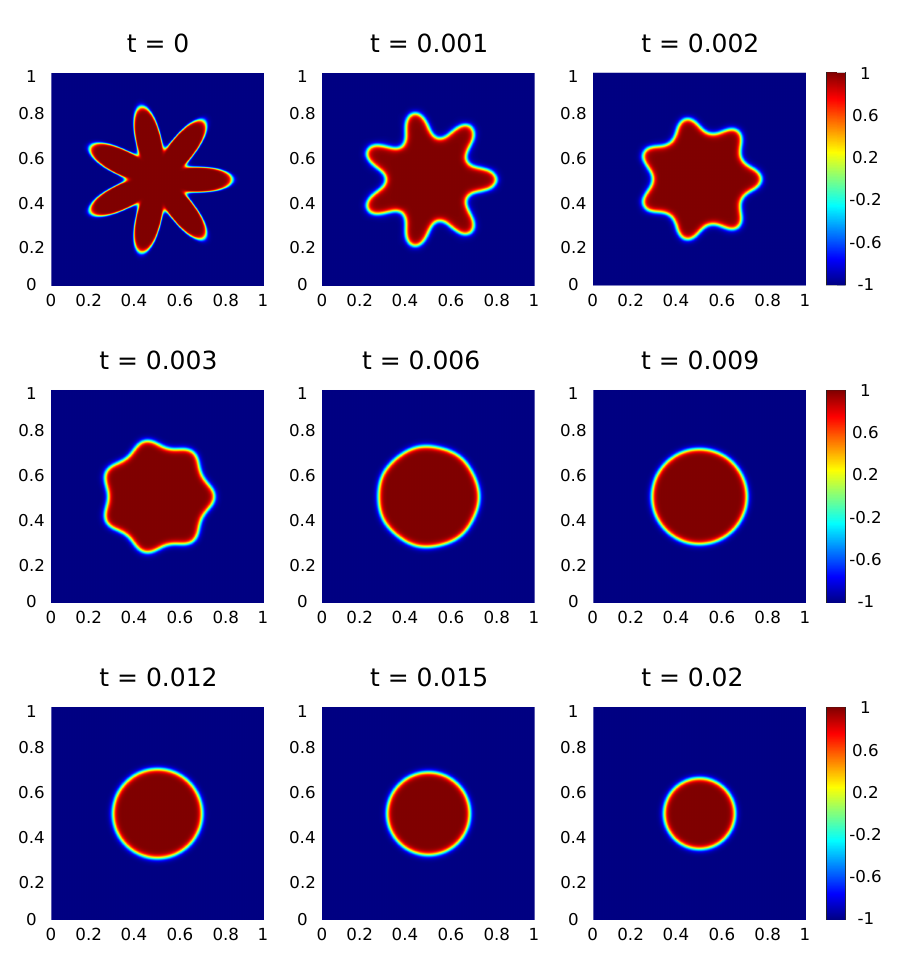}
    \caption{Solution evolution of the star-shaped interface at various time snapshots using the proposed SDMM approach.}
    \label{fig:ic3_sol}
\end{figure}

\begin{figure}[H]
    \centering
    \includegraphics{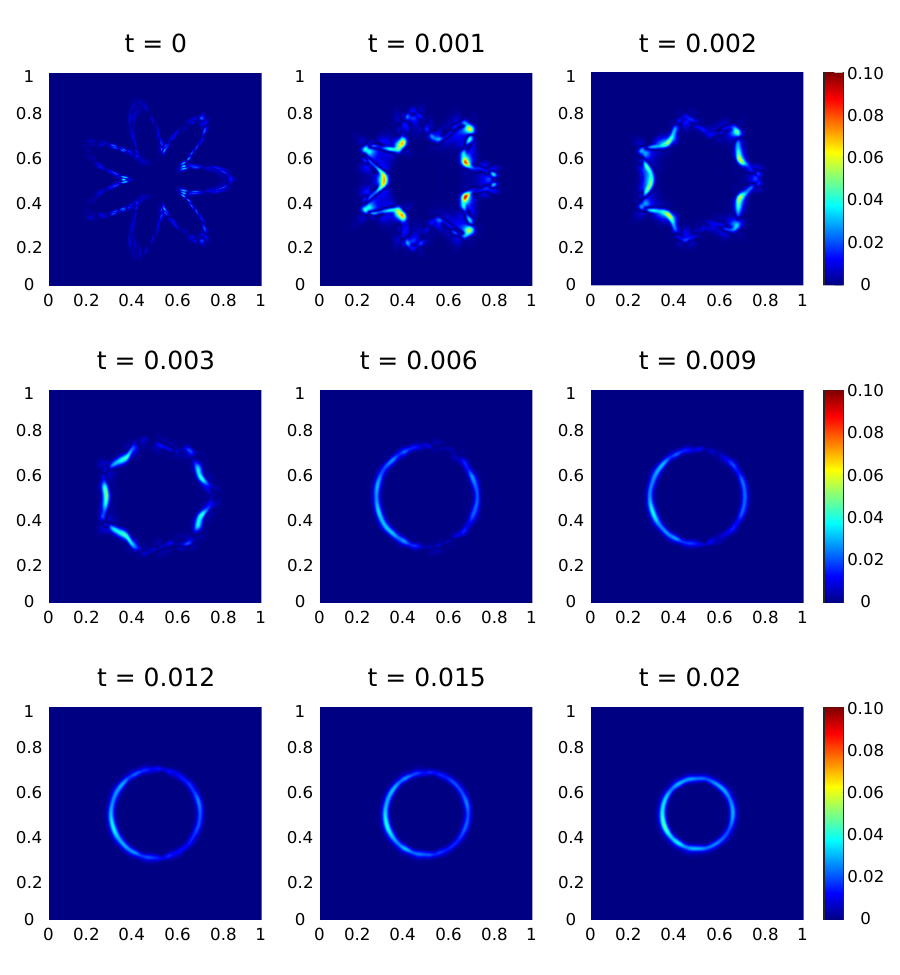}
    \caption{The absolute error between the SDMM and the reference solution.}
    \label{fig:ic3_err}
\end{figure}

\begin{figure}[H]
    \centering
    \includegraphics[scale=0.55]{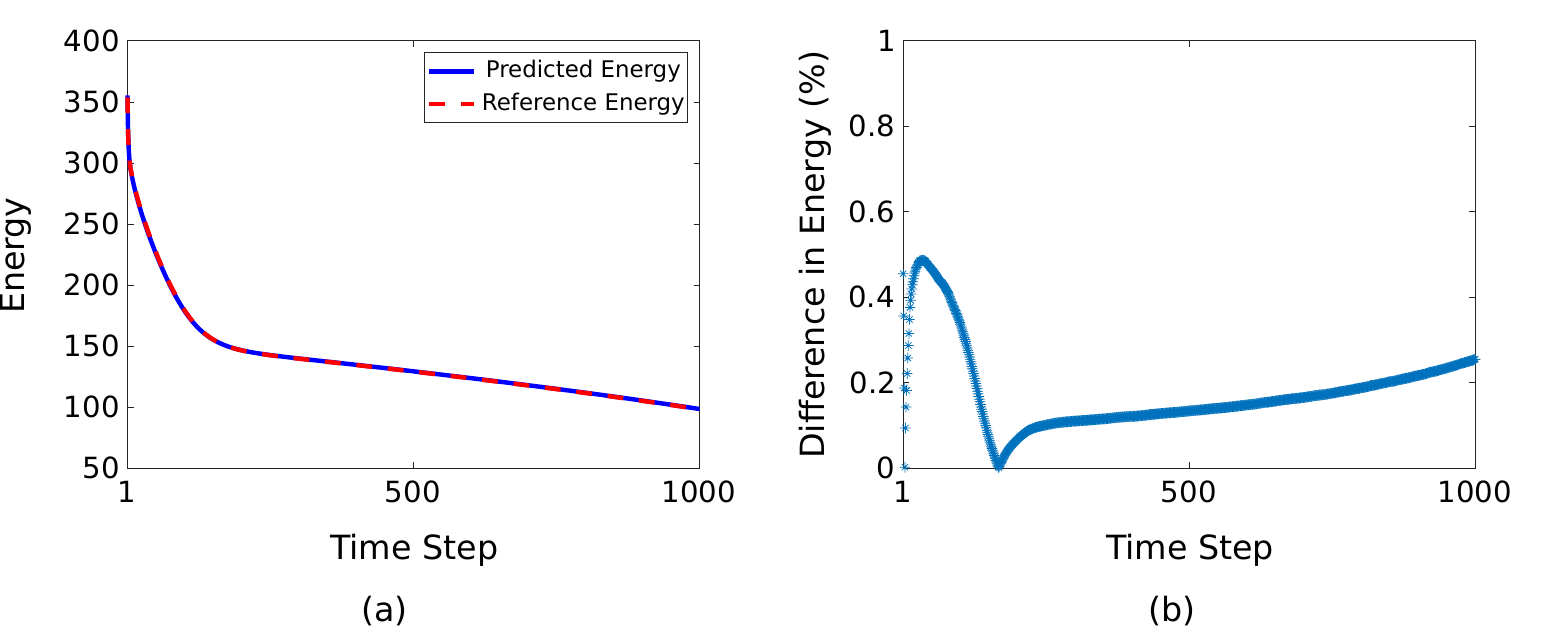}
    \caption{(a) Comparison of the energy between the solution obtained by the proposed SDMM method and the reference solution. (b) Difference in energy between the SDMM's prediction and reference solution (in \%).}
    \label{fig:ic3_energy_ref_pred}
\end{figure}

Figure~(\ref{fig:ic3_sol}) shows the evolution of the solution using the proposed SDMM method. As it can be seen that the system quickly evolves until $t = 0.003$ where the curvature decreases rapidly. This is also evidenced in the energy plot in figure~(\ref{fig:ic3_energy_ref_pred}), where the energy quickly decreases. Further, as evidenced in figure~(\ref{fig:ic3_err}) the absolute error between the SDMM predicted and reference solution is small thereby demonstrating that the the proposed method is highly accurate. \\

The proposed method is approximately 9 times faster than the finite element method (FEM) (shown in figure~(\ref{fig:ic3_refvsNN_time}), showcasing a substantial speed improvement. Furthermore, compared to the reference solution on a mesh size of $2048\times2048$, the root mean squared error of the predicted solution is $9.3992 \times 10^{-6}$, indicating a high degree of precision. The energy predicted by the proposed method also matches the finite element solutions remarkably well, as shown in figure~\ref{fig:ic3_energy_ref_pred}. 

\begin{figure}[H]
    \centering
    \includegraphics[scale=0.375]{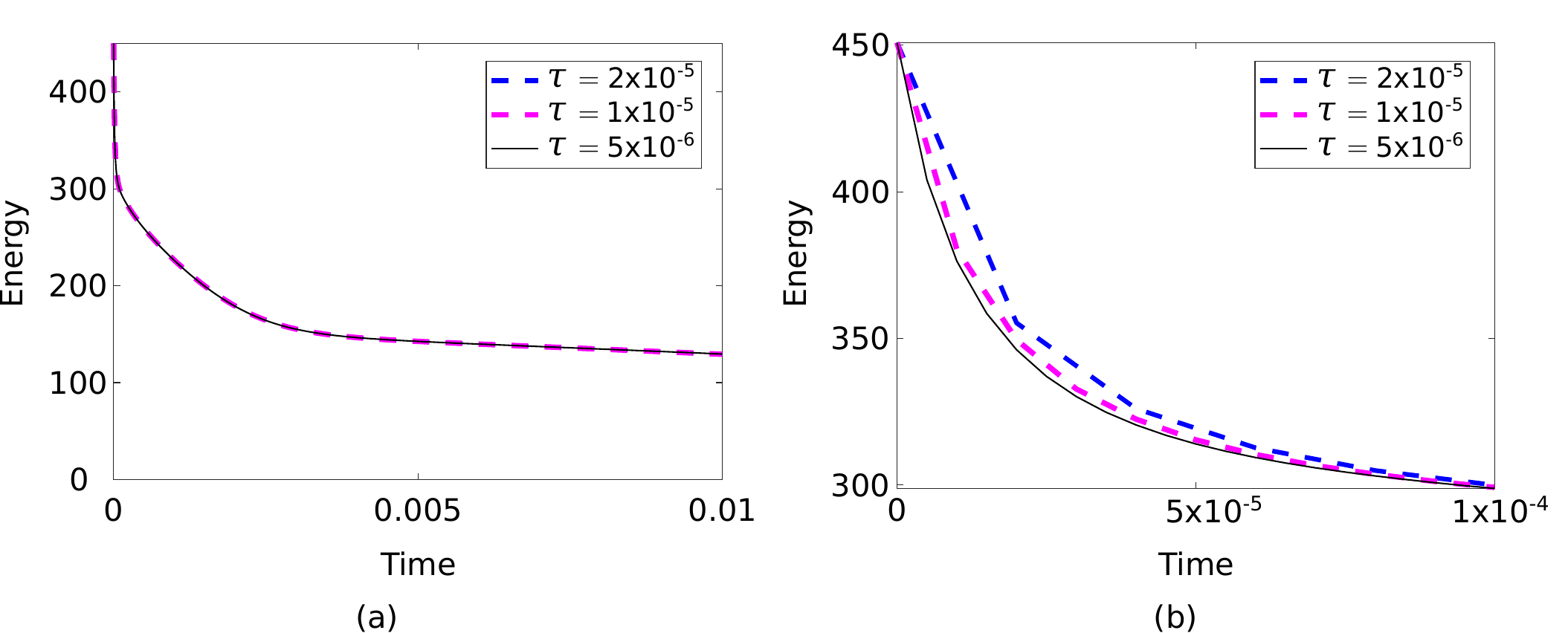}
    \caption{(a) Variation of energy with time using the proposed SDMM method for three different time steps (b) Magnified view of the variation in energy.}
    \label{fig:ic3_energy_time_convergence}
\end{figure}
To further investigate the convergence of the solution with time step $\tau$, the current problem is solved using different time steps, $\tau = 2 \times 10^{-5},\, 1\times 10^{-5},\, 5\times 10^{-6}$. The corresponding energy evolution and a magnified view for the three different time steps are shown in figure~\ref{fig:ic3_energy_time_convergence}. Figure~\ref{fig:ic3_energy_time_convergence}(a) shows that the predicted energy for all the three time steps matches closely. The magnified view in Figure~\ref{fig:ic3_energy_time_convergence}(b) for the early stages of the evolution of the phase field shows that the energy converges as the time step size is reduced. 

\subsection{Test 2: Coarsening Problem}\label{sec:coarsening}
For the second test case, a widely studied phase separation problem known as the coarsening problem exhibiting dynamic phase evolution is considered. The domain for the present system is chosen as $\Omega = [0,1]\times[0,1] \in \mathbb{R}^2$. The value of the time step is taken as $\tau = 5 \times 10^{-5}$ and simulated until a total time of $T = 0.04$. A rectangular grid mesh is chosen, which contains $2048^2$ elements.




\begin{figure}[H]
    \centering
    \includegraphics[scale=1]{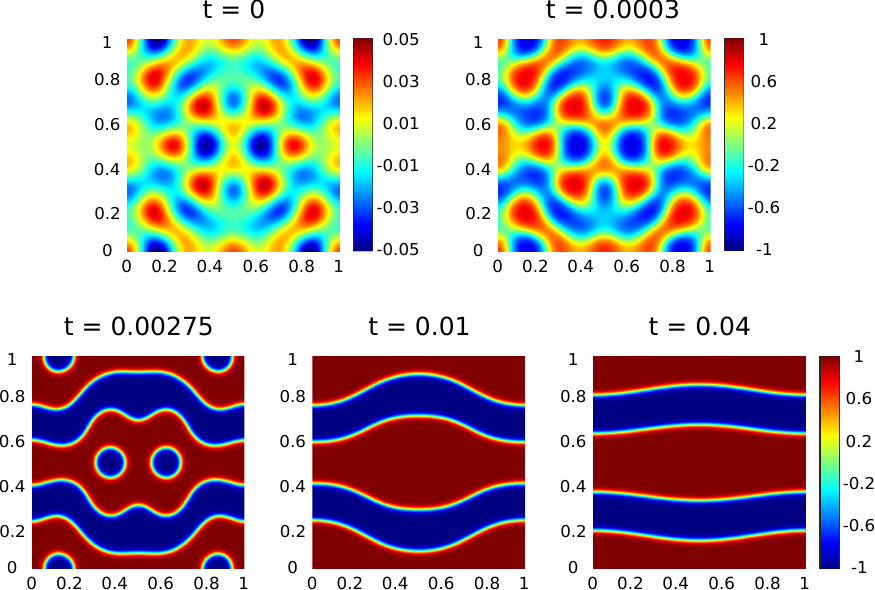}
    \caption{Solution evolution of the coarsening problem at various time snapshots using the proposed SDMM approach.}
    \label{fig:ic1_pred}
\end{figure}

\begin{figure}[H]
    \centering
    \includegraphics[scale=0.5]{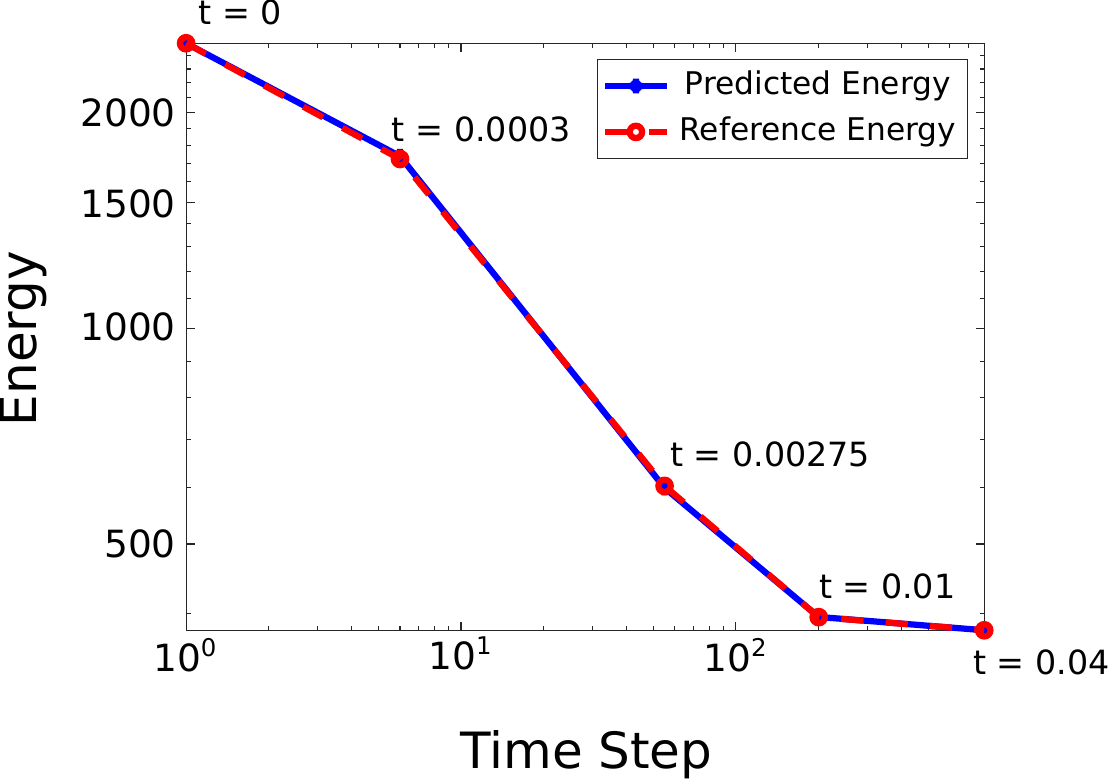}
    \caption{Variation of energy between the SDMM predicted solution and the reference solution is plotted over time for the coarsening problem.}
    \label{fig:ic1_energy}
\end{figure}

\begin{figure}[H]
    \centering
    \includegraphics[scale=1]{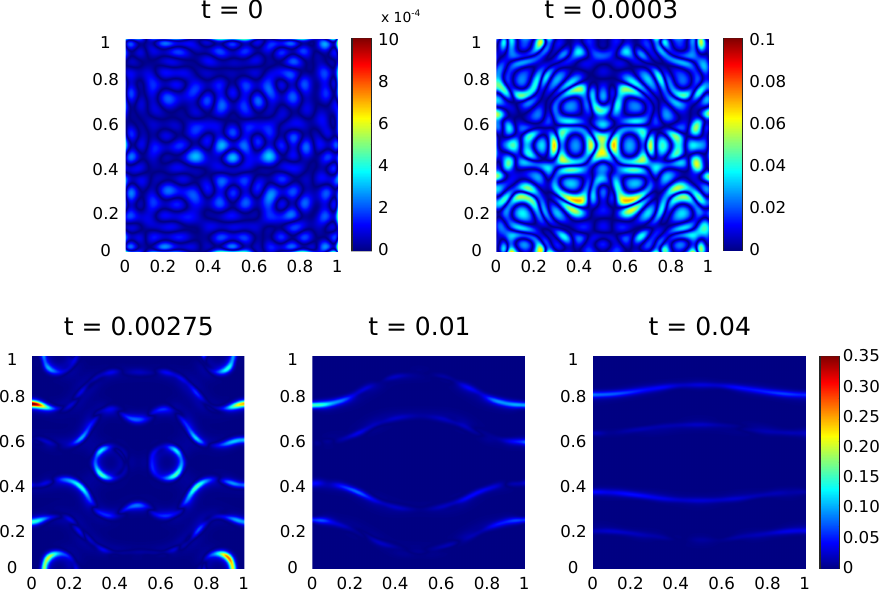}
    \caption{Absolute error ($\phi_{\text{abs-error}}$) between the SDMM predicted and reference solutions at various time snapshots for the coarsening problem.}
    \label{fig:ic1_err_ref_pred}
\end{figure}

\noindent Figure~(\ref{fig:ic1_pred}) and shows the evolution of the solution using the proposed SDMM approach. It's evident that the initial phases undergo rapid evolution before coalescing, a process that continues until, $t = 0.00275$. This coalescence is also reflected in the energy comparison shown in figure~(\ref{fig:ic1_energy}), where the energy initially decreases rapidly before the solution's evolution slows down, indicating the separation of the two distinct phases. 
In addition, figure~(\ref{fig:ic1_err_ref_pred}) shows that the absolute error is small, suggesting a good match between the proposed method and the finite element method. The accuracy of the proposed method is also evident from the good match in energy with the finite element method, as shown in figures~(\ref{fig:ic1_energy}). This validates the effectiveness and accuracy of our proposed method.

\begin{figure}[H]
    \centering
    \includegraphics[scale=0.825]{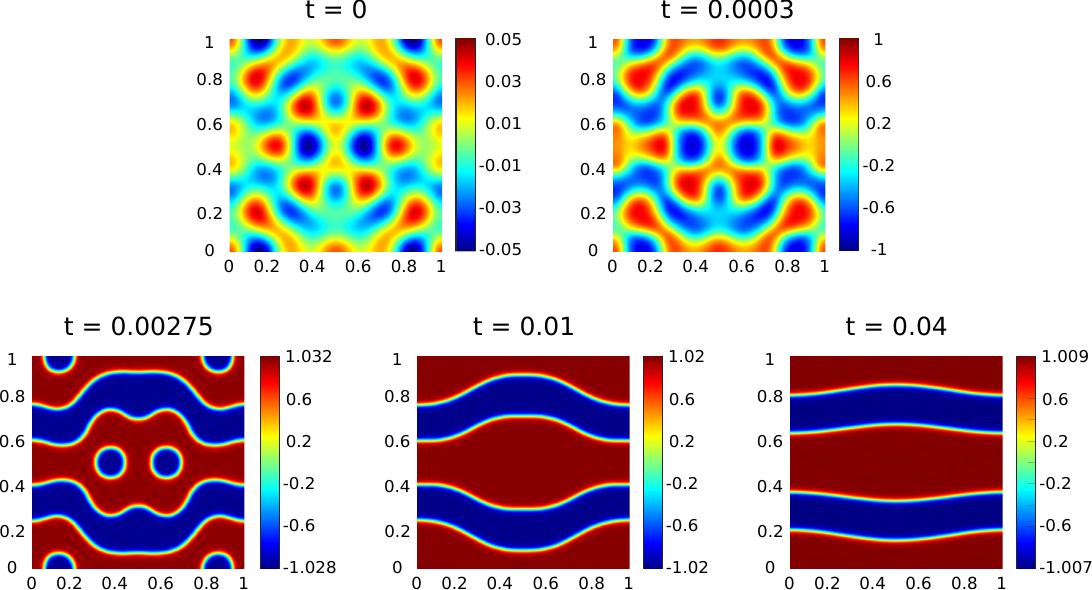}
    \caption{Solution evolution of the coarsening problem at various time snapshots using the proposed SDMM approach without the `$tanh$' transformation.}
    \label{fig:ic1_pred_noTanh}
\end{figure}

\begin{figure}[H]
    \centering
    \includegraphics[scale=0.425]{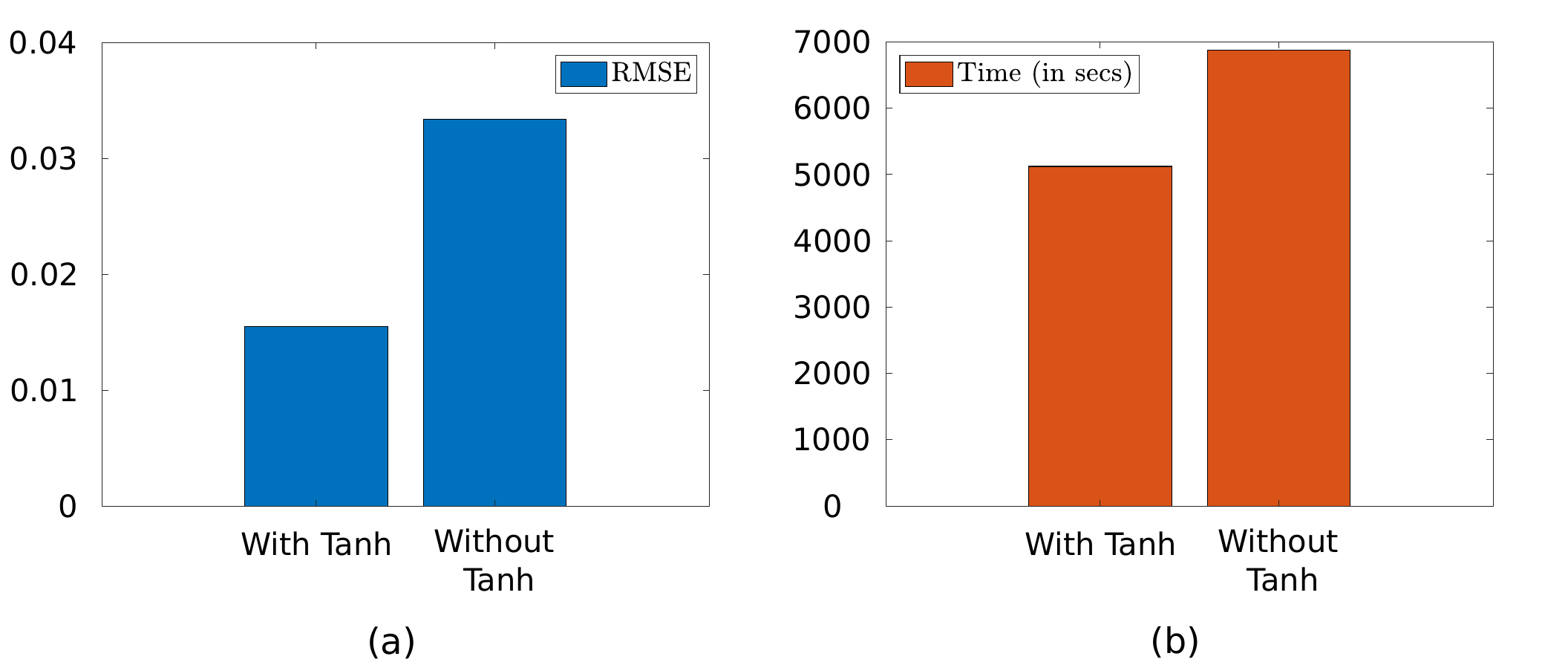}
    \caption{Bar graph showing the (a) variation of errors $\mathcal{E}_{\text{SDMM}}$ between the SDMM predicted solution and the reference solution and (b) the training time required with and without using the `$\tanh$' transformation using the proposed SDMM approach.}
    \label{fig:ic1_rmse_time_withwo_tanh}
\end{figure}

\noindent Furthermore, figure~(\ref{fig:ic1_pred_noTanh}) shows the solution predicted at various time steps using the SDMM method without any `$\tanh$' transformation where the solution is unbounded. A comparison of time and error metrics with and without the $\tanh$ transformation is presented in figure~(\ref{fig:ic1_rmse_time_withwo_tanh}). Clearly, the SDMM method with the $\tanh$ transformation exhibits significantly improved accuracy and speed compared to the SDMM method without any transformation, where the solution remains unbounded. 

\begin{figure}[H]
    \centering
    \includegraphics[scale=0.5]{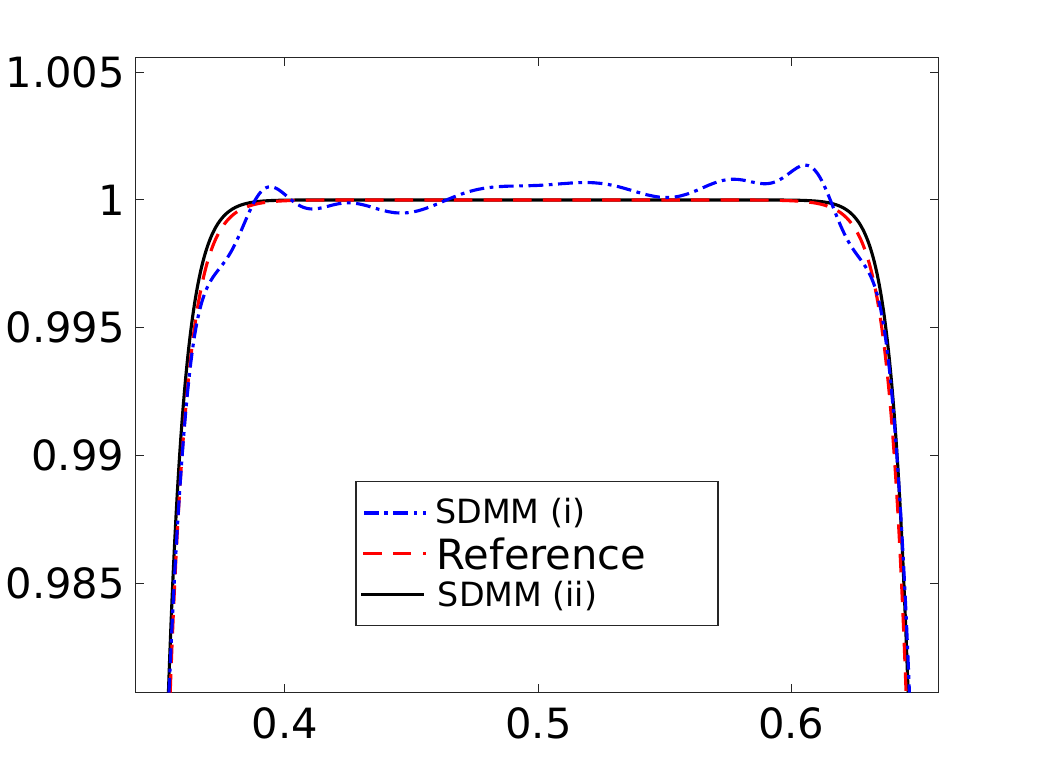}
    \caption{Comparison of $\phi$ between SDMM (\RN{1}), SDMM (\RN{2}), and the reference solution along a slice at $x = 0.5$ and $t=0.395$. Here, SDMM (\RN{1}) is the solution without the `$\tanh$' transformation and SDMM (\RN{2}) is with the `$\tanh$' transformation. In SDMM (\RN{1}) a larger neural network is considered compared to SDMM (\RN{2}). Further, in SDMM (\RN{1}) for every time step 100 LBFGS iterations have been used whereas SDMM (\RN{2}) is trained using 30 LBFGS iterations.}    \label{fig:ic1_ref_pred_step_790_withwoTanh}
\end{figure}

To achieve comparable accuracy levels between solutions obtained with and without a $\tanh$ transformation, we considered employing a larger network and increasing the LBFGS iterations for the latter approach. Figure~(\ref{fig:ic1_ref_pred_step_790_withwoTanh}) illustrates a comparison of solution predictions at $T = 0.0395$ among the reference solution, the $\tanh$-transformed SDMM solution, and the SDMM solution without any transformation. Notably, the $\tanh$-transformed SDMM solution closely approximates the reference solution, whereas the SDMM solution without any transformation still exhibits minor deviations from the phase boundary. In terms of computational efficiency, the $\tanh$-transformed SDMM (\RN{1}) solution in figure~(\ref{fig:ic1_ref_pred_step_790_withwoTanh}) offers a threefold speedup compared to SDMM (\RN{2}) without any transformation.

\begin{figure}[H]
    \centering
    \includegraphics[scale=0.325]{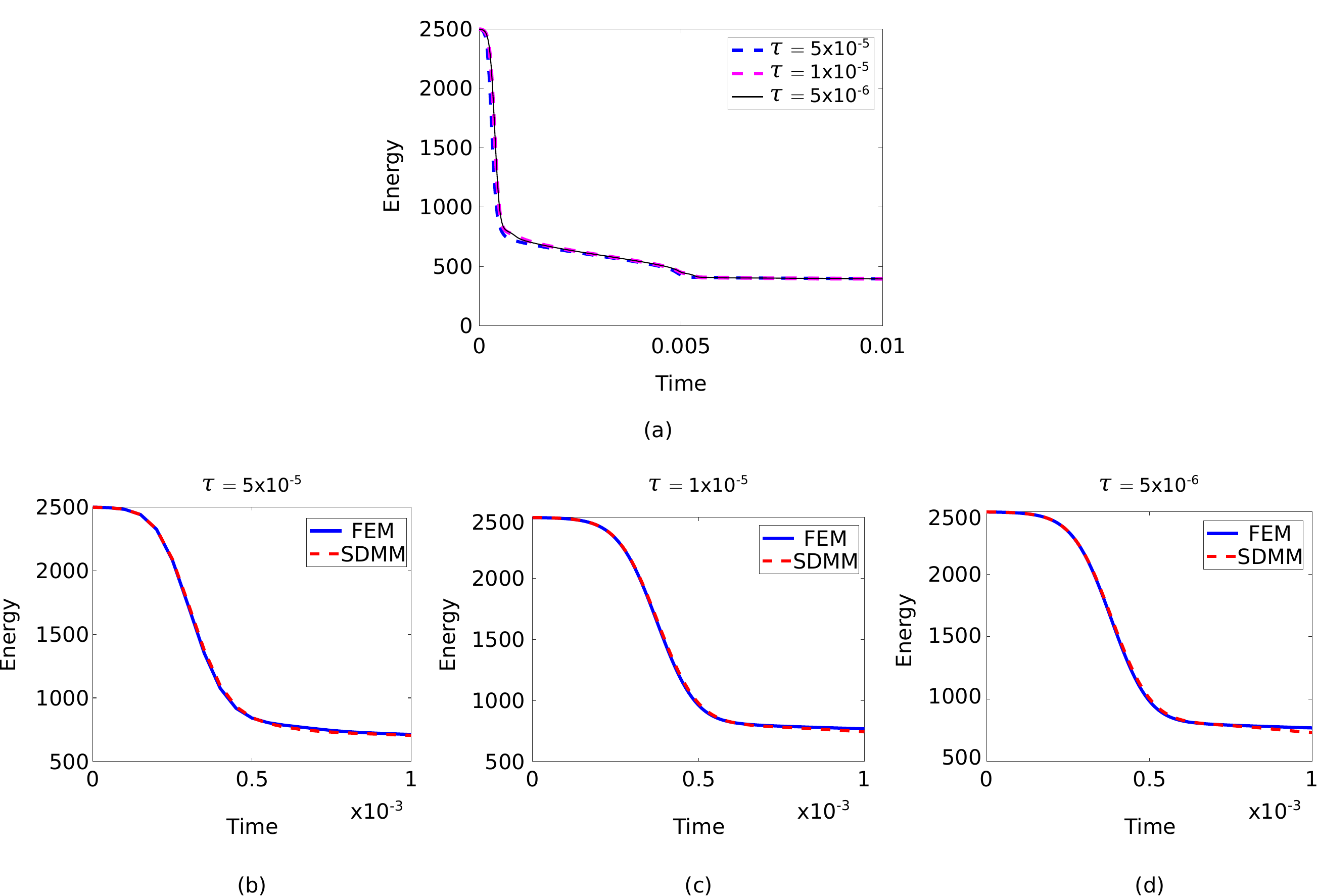}
    \caption{(a)Variation of energy with time using the proposed SDMM method for three different time steps. Comparison of energy predicted using the reference FEM solution and the proposed SDMM method for (b) $\tau = 5\times 10^{-5}$, (c) $\tau = 1\times 10^{-5}$, and (d) $\tau = 5\times 10^{-6}$.}
    \label{fig:ic1_energy_time_convergence}
\end{figure}

Finally, to investigate the convergence of the solution against time steps, the problem was solved using different time steps: $\tau = 5 \times 10^{-5},\, 1 \times 10^{-5},\, 5 \times 10^{-6}$. The corresponding energy evolution is shown in figure~(\ref{fig:ic1_energy_time_convergence})(a), which shows the predicted energy is almost the same for all of the three time steps. 
Further, it can be seen in figure~(\ref{fig:ic1_energy_time_convergence})(b-d) that even at the early stages where the evolution is much rapid, the energy predicted by the SDMM and the finite element method matches closely for each of the time step used. 

\subsection{Test 3: Random Initial Condition}\label{sec:RandomIC}
\begin{figure}[b!]
    \centering
    \includegraphics[scale=0.425]{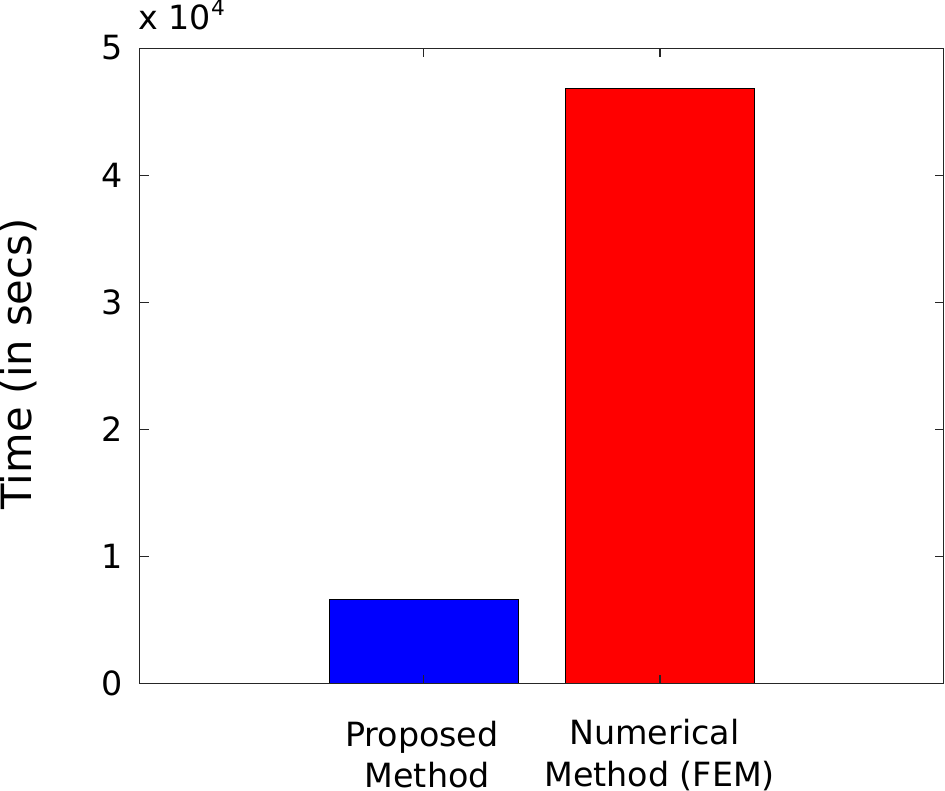}
    \caption{Bar graph comparing the computational wall time for the proposed SDMM method using GPU and the numerical method employing 24 CPU cores. Details of the CPU and GPU systems used for the simulations are given in \ref{sec:comp_resources}.}
    \label{fig:ic4_refvsNN_time}
\end{figure}

As a final test case, a random initial condition has been chosen so that a complex evolution of phases can be observed. The domain for the system is chosen as $\Omega  = [0,1] \times [0,1] \in \mathbb{R}^2$. The time step $\tau = 2 \times 10^{-5}$ is chosen and simulated until a total time of $T = 0.02$. The domain is meshed with $2048^2$ quadrilateral elements. A no-flux boundary condition ($\grad \phi = 0$) has been chosen. The numerical solution has been generated using FEniCS on a $2048 \times 2048$ mesh. The total wall time for the numerical simulation on a 24-core CPU was 46,800 seconds, whereas the solution using the proposed method took 6,600 seconds on a single GPU, achieving approximately a 7x speedup compared to the numerical method (shown in figure~(\ref{fig:ic4_refvsNN_time}).

\begin{figure}[H]
    \centering
    \includegraphics{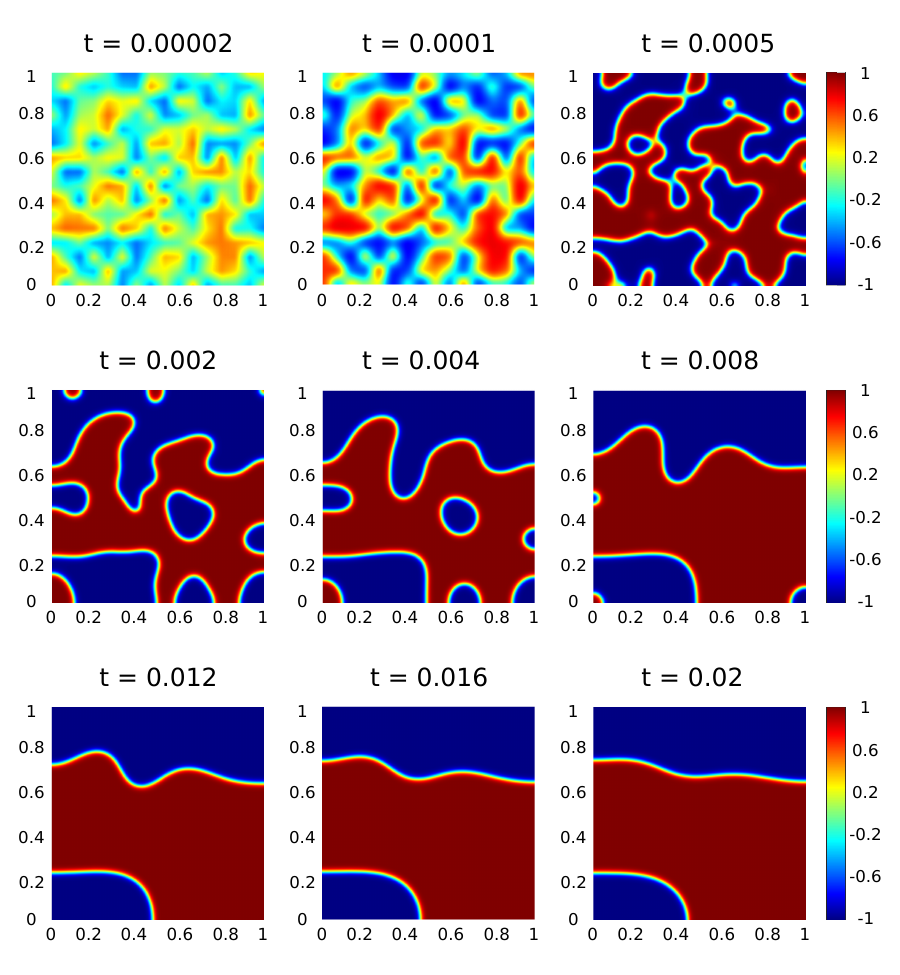}
    \caption{Phase evolution for the  problem with random initial condition at various time snapshots using the proposed SDMM approach with the `$tanh$' transformation.}
    \label{fig:ic4_pred_Tanh}
\end{figure}

\noindent Figure~(\ref{fig:ic4_pred_Tanh}) shows the phase evolution predicted by the proposed SDMM method for a random initial condition. Initially, at $t=0.0001$, the system begins in a near-uniform state. As time progresses, phase separation becomes evident, with distinct regions of differing phase values emerging. By $t = 0.002$, the system shows more pronounced phase boundaries, indicating the growth and coarsening of these regions. The evolution continues, with phase domains becoming more defined and stable as time advances. By the final time step $t=0.02$, the system approaches a more stable configuration, with clear, well-separated phases indicating the system's progression towards equilibrium. \\

\begin{figure}[H]
    \centering
    \includegraphics{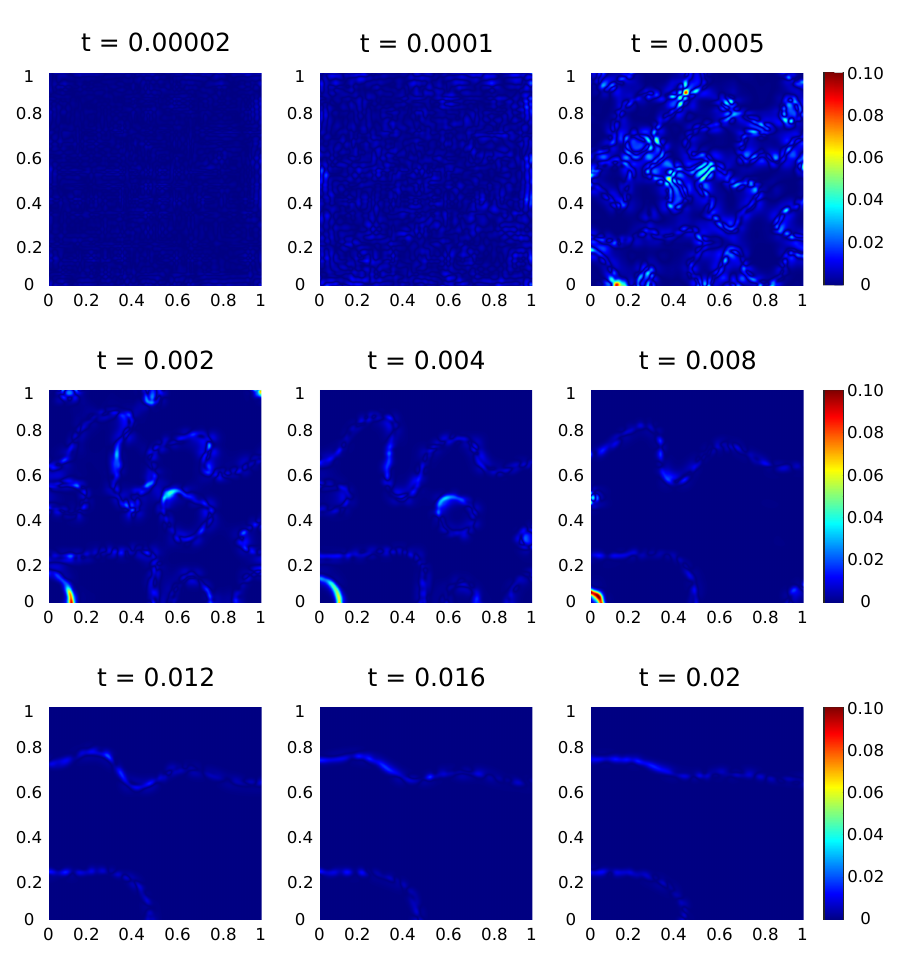}
    \caption{Absolute error $(\phi_{abs-error})$ in the phase value between the SDMM predicted and reference solution for the  problem with random initial condition at various time snapshots.}
    \label{fig:ic4_err_ref_pred}
\end{figure}

\noindent Figure~(\ref{fig:ic4_err_ref_pred})  presents the errors in the absolute values of the phases at specific time snapshots. The total error $\mathcal{E}_{\text{SDMM}}$ is $3.5 \times 10^{-3}$ and highlights the high accuracy of the proposed SDMM method.

\begin{figure}[t!]
    \includegraphics[scale=0.425]{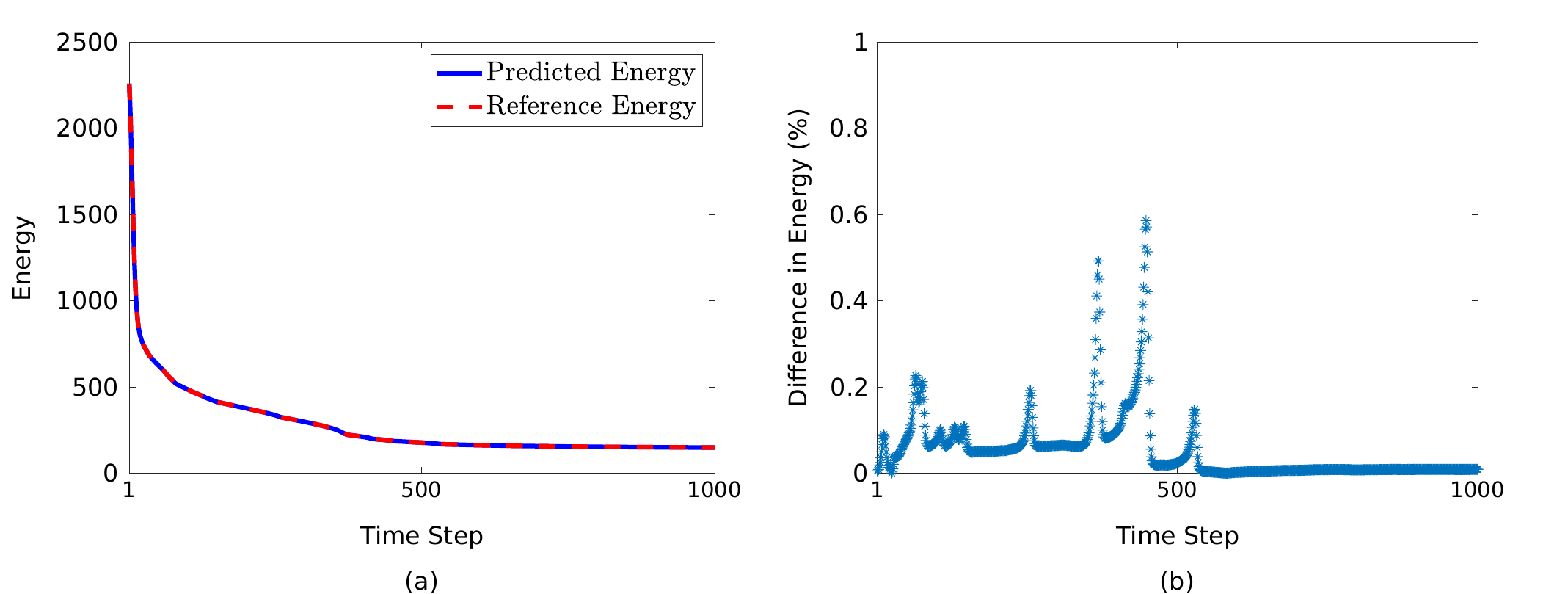}
    \caption{(a) Comparison of the energy between the solution obtained by the proposed SDMM method and the reference solution. (b) Difference in energy between the SDMM's prediction and reference solution (in \%).}
    \label{fig:ic4_energy_ref_pred}
\end{figure}

\noindent Finally, figure (\ref{fig:ic4_energy_ref_pred}) demonstrates a remarkable agreement between the energy from the proposed method and the reference solution. The maximum difference between the energies is approximately $0.6\%$ indicating a high level of accuracy.

\section{Conclusions}\label{sec:conclusion}
A separable neural network approximation of the phase field is used in a minimizing movement scheme to solve the $L^2$ gradient flow of the Ginzburg Landau free energy functional, which yields the solution of Allen Cahn equation. The key aspects of the proposed method are mentioned below. 
Firstly, the gradient flow approach uses an energy functional having lower order derivatives of the phase field than the strong form of the Allen-Cahn equation. This allows for bypassing the high computational cost associated with the higher order derivative calculation.
Secondly, the adoption of minimizing movement scheme allows for obtaining the solution by minimizing the functional at each time step, eliminating the computational complexity of a space-time approach. 
Thirdly, The separable neural network represents the phase field using low-rank tensor decomposition, enabling the use of forward-mode automatic differentiation for fast derivative calculations.   
Fourthly, the framework allows for accurate calculation of the energy functional through Gauss quadrature overcoming the inaccuracies of the collocation approach. 
Fifthly, the proposed hyperbolic tangent transformation applied to the neural network's output constrains the solution within the two-phase values, enhancing both the accuracy and efficiency of the method. Additionally, we provide a theoretical guarantee that the transformation preserves the unconditional energy stability of the minimizing movement scheme. 

State-of-the-art machine learning methods for solving the Allen Cahn equation is erroneous for longer time range. The proposed method addresses the limitation by providing highly accurate solution for long time range. 
It demonstrates remarkable computational efficiency, achieving an order of magnitude improvement over the finite element method. The proposed approach is versatile and should be applicable to other complex gradient flow problems. 

\section*{Acknowledgements}
We acknowledge the financial support by grant DE-SC0023432 funded by the U.S. Department of Energy, Office of Science. This research used resources of the Deep Blizzard GPU cluster at MTU and the National Energy Research Scientific Computing Center, a DOE Office of Science User Facility supported by the Office of Science of the U.S. Department of Energy under Contract No. DE-AC02-05CH11231, using NERSC awards BESERCAP0025205 and BES-ERCAP0025168.

\newpage
\clearpage
\newpage
\bibliography{main}

\newpage
\appendix
\section{Automatic Differentiation}\label{sec:AD}
Automatic differentiation (AD), often called algorithmic differentiation, is a widely used technique in machine learning for computing gradients and Hessians necessary for various optimization algorithms. Fundamentally, AD is an algorithmic approach where the computation of derivatives is decomposed into a sequence of basic operations, including addition, multiplication, and the derivatives of elementary functions like trigonometric and polynomial functions. There are two primary modes of automatic differentiation: forward mode and reverse mode. In forward mode, derivatives are calculated during the forward pass of the neural network. In reverse mode, the neural network function is initially evaluated during the forward pass, and then the derivatives are computed by working backward through the computational graph. To illustrate the difference a standard example from \cite{baydin2018automatic} is shown below. \\
\noindent Consider a function, $f(x_1,x_2) = \ln{(x_1)} - x_1 x_{2}^{2}$,

\begin{table}[H]
    \centering
    \begin{tabular}{cccc}
        \toprule
         & Forward primal trace & Forward tangent trace & Backward adjoint trace \\
         \tikzmark{a} & $v_{-1} = x_1$ & \tikzmark{aa} $\Dot{v}_{-1} = \Dot{x}_1 =  1$  & \tikzmark{bbb}$\bar{x}_1 = \bar{v}_{-1}$ \\
         & $v_0 = x_2$ & $\Dot{v}_{0} = \Dot{x}_2 =  0$  & $\bar{x}_2 = \bar{v}_0$ \\
         \hline
         & $v_1 = \ln{(v_{-1})}$ & $\Dot{v}_1 = \Dot{v}_{-1}/v_{-1}$  & $\bar{v}_{-1} = \bar{v}_{-1} + \bar{v}_1\frac{\partial\bar{v}_{1}}{\partial\bar{v}_{-1}}$ \\
         
         & $v_2 = v_{-1} v_0^2$ & $\Dot{v}_2 = 2 v_{-1} v_0 \Dot{v}_0 + \Dot{v}_{-1} v_0^2 $  & $\bar{v}_{-1} = \bar{v}_2\frac{\partial v_2}{\partial v_{-1}}$ \\
         
         & & & $\bar{v}_0 = \bar{v}_2\frac{\partial v_2}{\partial v_0}$ \\
         
         & $v_3 = v_1 + v_2$ & $\Dot{v}_3 = \Dot{v}_1 + \Dot{v}_2 $  & $\bar{v}_1 = \bar{v}_3 \frac{\partial v_3}{\partial v_2}$ \\
         & & & $\bar{v}_2 = \bar{v}_3 \frac{\partial v_3}{\partial v_2}$ \\ \\
         \hline
         \tikzmark{b} & $y= v_3$ & \tikzmark{bb} $\Dot{y} = \Dot{v}_3$  & \tikzmark{aaa}$\bar{v}_3 = \bar{y}$ \\ 
         \bottomrule
    \end{tabular}
    \caption{Forward and reverse mode AD for the example function $f(x_1,x_2)$. $v_{-1}, v_0$ denote the input variables, $v_k$ and $\Dot{v}_k$ are the primals and tangents evaluated during the forward pass, $\bar{v}_k$ are the adjoints computed during the backward pass. }
    \label{tab:AD}
    \begin{tikzpicture}[overlay,remember picture]
    \draw[->,line width=2pt, color=black] (pic cs:a) -- (pic cs:b);
    \end{tikzpicture}
    \begin{tikzpicture}[overlay,remember picture]
    \draw[->,line width=2pt, color=black] ($(pic cs:aa)+(-29pt,0ex)$) -- ($(pic cs:bb)+(-45pt,0ex)$);
    \end{tikzpicture}
    \begin{tikzpicture}[overlay,remember picture]
    \draw[->,line width=2pt, color=black] ($(pic cs:aaa)+(-40pt,0ex)$) -- ($(pic cs:bbb)+(-35pt,0ex)$);
    \end{tikzpicture}
\end{table}

\noindent In table~(\ref{tab:AD}), the first column shows how the primals $v_i$ (intermediate values) are computed during the forward pass. To compute the derivative of output $y$ with respect to $x_1$, during the forward pass, each intermediate variable $v_i$ is associated with a derivative $\Dot{v}_i = \frac{\partial v_i}{\partial x_1}$. On the other hand, reverse mode AD aligns with a standard backpropagation algorithm. In this approach, the derivatives are calculated in reverse by accumulating the adjoints $\bar{v}_k = \frac{\partial y}{\partial v_i}$ starting from a specific output. \\

\noindent In general for a function $h:\mathbb{R}^n \to \mathbb{R}^m$ with $n$ independent variables $x_i$ and $m$ dependent variables $y_j$, one forward pass would compute the derivatives $\frac{\partial y_j}{\partial x_i}$ for all $j=1,\cdots,m$. This is equivalent to computing one full column of the jacobian matrix $\mathbb{J}$. In contrast, in reverse mode AD during one backward pass would compute the derivatives $\frac{\partial y_j}{\partial x_i}$ for all $i=1,\cdots,n$, which is equivalent to computing one full row of the jacobian matrix $\mathbb{J}$. 
\begin{equation}
    \mathbb{J} = \begin{pmatrix}
    \frac{\partial y_1}{\partial x_1} & \cdots  & \frac{\partial y_1}{\partial x_n} \\
    \vdots & \ddots & \vdots \\
    \frac{\partial y_m}{\partial x_1} & \cdots  & \frac{\partial y_m}{\partial x_n}
\end{pmatrix}
\end{equation}

\noindent To summarize, in forward mode AD, the complete jacobian requires $n$ evaluations, while in reverse mode AD, the entire jacobian can be computed in $m$ evaluations. Therefore, when ($m>>n$) forward mode AD proves more efficient; conversely, when ($n>>m$) reverse mode AD is better suited.

\section{Separated and Non-Separated Functions}\label{sec:sep_vs_nonsep}
In the context of computational methods for solving boundary value problems and representing tensor fields on a structured grid, low-rank tensor decomposition can be an efficient technique. This approach involves breaking down a tensor field into simpler components, typically rank-1 tensors or vectors. The tensor field can then be reconstructed by taking the outer products of these rank-1 tensors. \\

\begin{figure}[!htb]
     \centering
     \subfloat[Non-Separated\\ approach]{\label{fig:non-separated}\includegraphics[scale=0.425]{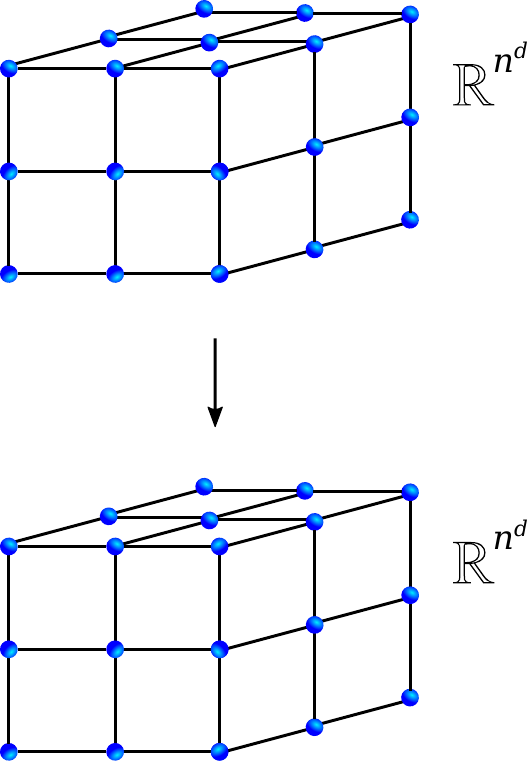}} \qquad
     \subfloat[Separated approach]{\label{fig:separated}\includegraphics[scale=0.425]{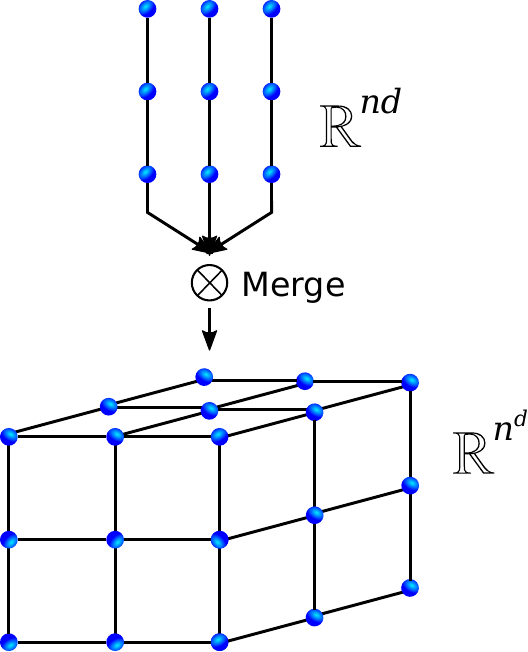}} \quad
     \caption{(a) Non-separated approach that requires $\mathcal{O}(n^d)$ evaluations for computing the $\mathbb{J}$ (Jacobian matrix) in both forward and reverse mode AD for a mapping from $\mathbb{R}^{n^{d}} \to \mathbb{R}^{n^{d}}$ (b) Separated approach (mapping from $\mathbb{R}^{nd} \to \mathbb{R}^{n^{d}}$) requires $\mathcal{O}(nd)$ evaluations for computing the $\mathbb{J}$ (Jacobian matrix) using forward mode AD and $\mathcal{O}(n^d)$ evaluations using reverse mode AD \cite{cho2024separable}}
     \label{fig:sep_vs_non-sep}
\end{figure}

\noindent For a given boundary value problem in $d$ dimensions, a low-rank tensor approximation method allows us to express the solution field as a tensor product of $d$ individual rank-1 tensors. This approach is referred to as a separated approach \cite{cho2024separable}. In a separated approach, predicting the solution field in a grid containing $N^d$ points requires only $N d$ collocation points as each dimension is treated separately, reducing the overall computational complexity. In contrast, a non-separated approach would necessitate using all $N^d$ collocation points in the grid to predict the solution field. Therefore, when predicting the solution on a structured grid a separated approach significantly reduces the number of collocation points compared to a non-separated approach. Especially for higher-dimensional problems a separated approach can be much more computationally efficient than a non-separated approach.  Figure~(\ref{fig:sep_vs_non-sep}), shows a schematic representation of the separated and non-separated approaches and highlights the difference in their computational requirements.
In the subsequent section, a separable neural network approach is described which leverages the principles of low-rank tensor decomposition for predicting a tensor field.

\section{Details of the Computational Resources}\label{sec:comp_resources}
Nvidia A100 GPU (6912 CUDA cores, 432 Tensor cores, and 40 GB of HBM2 vRAM) is used for training the neural networks. For inferencing and generating the reference solutions using FENICS, a Dell precision 3660 workstation with Intel core i9-9700k containing 32 cores (5.6 GHz Turbo) and 64 GB RAM has been utilized. The software packages used for all the computations are PyTorch 2.0.1 and MATLAB R2023b.

\end{document}